\documentclass[titlepage]{article}

\usepackage{hyperref} 
\hypersetup{
      breaklinks=true,  
      colorlinks=true,
      urlcolor=blue,
      linkcolor=blue,
      citecolor=blue,
}
\usepackage{authblk}

\usepackage[T1]{fontenc}
\usepackage{ae}
\usepackage{aecompl}
\usepackage[utf8]{inputenc}
\usepackage[ngerman,english]{babel}
\usepackage{natbib}
\usepackage[export]{adjustbox} 
\usepackage{color} 
\usepackage{enumerate} 
\usepackage[left=1.25in,right=1.25in]{geometry} 
\usepackage{longtable} 
\usepackage{arydshln} 
\usepackage{rotating} 
\usepackage{pdflscape} 
\usepackage{amsmath}
\usepackage{mathtools}
\usepackage{amssymb}
\usepackage{pbox} 
\usepackage{booktabs}
\usepackage{fancyvrb} 
\usepackage{grffile} 
\usepackage{float}
\usepackage{subcaption}
\usepackage{setspace} 
\usepackage{multirow} 
\usepackage{fancyhdr}
\usepackage{abstract}

\usepackage{tikz}
\usetikzlibrary{positioning}
\tikzset{every picture/.style={font=\sffamily}}

\newcounter{result}[section]
\newenvironment{result}[1][]{\refstepcounter{result}\par\medskip \noindent
\textbf{Result \theresult} \rmfamily \itshape}{\medskip}
  
\title{\rule{\textwidth}{1pt}\\\LARGE{\textbf{Deep learning with convolutional neural networks for brain mapping and decoding of movement-related information from the human EEG}}\\\rule{\textwidth}{1pt}\\\
\\ \large{Short title: \itshape{Convolutional neural networks in EEG analysis}} \\
\vspace{0.5cm} 
\normalsize{Keywords: \itshape{Electroencephalography, EEG analysis, machine learning, end-to-end learning,
brain-machine interface (BCI), brain-computer interface (BMI), model interpretability, brain mapping}}}

\author[a,b]{Robin Tibor Schirrmeister }
\author[c,b]{Jost Tobias Springenberg}
\author[a,b,d]{Lukas Dominique Josef Fiederer}
\author[a,b]{Martin Glasstetter}
\author[e,b]{Katharina Eggensperger}
\author[f,b]{Michael Tangermann}
\author[e,b]{Frank Hutter}
\author[g,b]{Wolfram Burgard}
\author[a,b]{Tonio Ball\thanks{Corresponding author: tonio.ball@uniklinik-freiburg.de}}
\affil[a]{Intracranial EEG and Brain Imaging lab, Epilepsy Center, Medical Center – University of Freiburg}
\affil[b]{BrainLinks-BrainTools Cluster of Excellence, University of Freiburg}
\affil[c]{Machine Learning Lab, Computer Science Dept., University of Freiburg}
\affil[d]{Neurobiology and Biophysics, Faculty of Biology, University of Freiburg}
\affil[e]{Machine Learning for Automated Algorithm Design Lab, Computer Science Dept., University of Freiburg}
\affil[f]{Brain State Decoding Lab, Computer Science Dept., University of Freiburg}
\affil[g]{Autonomous Intelligent Systems Lab, Computer Science Dept., University of Freiburg}

\newcommand{\beginsupplement}{%
        \setcounter{table}{0}
        \renewcommand{\thetable}{S\arabic{table}}%
        \setcounter{figure}{0}
        \renewcommand{\thefigure}{S\arabic{figure}}%
     }
\begin{document}

\maketitle


\begin{abstract}
Deep learning with convolutional neural networks (deep ConvNets) has revolutionized computer vision through end-to-end learning, i.e. learning from the raw data.
Now, there is increasing interest in using deep ConvNets for end-to-end EEG analysis.
However, little is known about many important aspects of how to design and train ConvNets for end-to-end EEG decoding, and there is still a lack of techniques to visualize the informative EEG features the ConvNets learn. 

Here, we studied deep ConvNets with a range of different architectures, designed for decoding imagined or executed movements from raw EEG.
Our results show that recent advances from the machine learning field, including batch normalization and exponential linear units, together with a cropped training strategy, boosted the deep ConvNets decoding performance, reaching or surpassing that of the widely-used filter bank common spatial patterns (FBCSP) decoding algorithm.
While FBCSP is designed to use spectral power modulations, the features used by ConvNets are not fixed a priori.
Our novel methods for visualizing the learned features demonstrated that ConvNets indeed learned to use spectral power modulations in the alpha, beta and high gamma frequencies.
These methods also proved useful as a technique for spatially mapping the learned features, revealing the topography of the causal contributions of features in different frequency bands to decoding the movement classes. 

Our study thus shows how to design and train ConvNets to decode movement-related information from the raw EEG without handcrafted features and highlights the potential of deep ConvNets combined with advanced visualization techniques for EEG-based brain mapping.

\end{abstract}

\section{Introduction}\label{sec:introduction}

Machine-learning techniques allow to extract information from electroencephalographic (EEG) recordings of brain 
activity and therefore play a crucial role in several important EEG-based research and application areas.
For example, machine-learning techniques are a central component of many EEG-based brain-computer interface (BCI) 
systems for clinical applications.
Such systems already allowed, for example, persons with severe paralysis to communicate 
\citep{nijboer_p300-based_2008}, to draw pictures \citep{munsinger_brain_2010},
and to control telepresence robots \citep{tonin_brain-controlled_2011}. 
Such systems may also facilitate stroke rehabilitation \citep{ramos-murguialday_brainmachine_2013} and may be used in 
the treatment of epilepsy \citep{gadhoumi_seizure_2016} (for more examples of potential clinical applications, see 
\citet{moghimi_review_2013}).
Furthermore, machine-learning techniques for the analysis
of brain signals, including the EEG, are increasingly recognized as
novel tools for neuroscientific inquiry 
\citep{das_predicting_2010,knops_recruitment_2009,kurth-nelson_fast_2016,stansbury_natural_2013}.

However, despite many examples of impressive progress, there is still
room for considerable improvement with respect to the accuracy of
information extraction from the EEG and hence, an interest in
transferring advances from the area of machine learning to the field of
EEG decoding and BCI. A recent, prominent example of such an
advance in machine learning is the application of convolutional neural
networks (ConvNets), particularly in computer vision tasks. Thus, first
studies have started to investigate the potential of ConvNets for
brain-signal decoding (\citep{
antoniades_deep_2016,
bashivan_learning_2016,
cecotti_convolutional_2011,
hajinoroozi_eeg-based_2016,
lawhern_eegnet:_2016,
liang_predicting_2016,
manor_multimodal_2016,
manor_convolutional_2015,
page_wearable_2016,
ren_convolutional_2014,
sakhavi_parallel_2015,
george_single-trial_2016,
stober_learning_2016,
stober_using_2014,
sun_remembered_2016,
tabar_novel_2017,
tang_single-trial_2017,
thodoroff_learning_2016,
wang_deep_2013}, see Supplementary Section \ref{supp:subsec-related-work} for more details on these studies).
Still, several important methodological questions on EEG analysis with ConvNets remain, as 
detailed below and addressed in the present study.

ConvNets are artificial neural networks that can learn local patterns in
data by using convolutions as their key component (also see Section 
\ref{subsec:convnets}).
ConvNets vary in the number of convolutional layers, ranging from shallow architectures with just one convolutional 
layer such as in a successful speech recognition ConvNet \citep{abdel-hamid_convolutional_2014}
over deep ConvNets with multiple consecutive convolutional layers \citep{krizhevsky_imagenet_2012}
to very deep architectures with more than 1000 layers as in the case of the recently developed residual networks
\citep{he_deep_2015}.
Deep ConvNets can first extract local, low-level features
from the raw input and then increasingly more global and high
level features in deeper layers. For example, deep ConvNets can learn to
detect increasingly complex visual features (e.g., edges, simple shapes,
complete objects) from raw images. Over the past years, deep
ConvNets have become highly successful in many application areas, such
as in computer vision and speech recognition, often outperforming
previous state-of-the-art methods (we refer to \citet{lecun_deep_2015} and \citet{schmidhuber_deep_2015} for recent 
reviews).
For example, deep ConvNets reduced the error rates on the ImageNet
image-recognition challenge, where 1.2 million images must be classified
into 1000 different classes, from above 26\% to below 4\% within 4 years
\citep{he_deep_2015,krizhevsky_imagenet_2012}.
ConvNets also reduced error rates in recognizing speech,
e.g., from English news broadcasts
\citep{sainath_deep_2015,sercu_very_2016}; however, in this field, hybrid models
combining ConvNets with other machine-learning components, notably
recurrent networks, and deep neural networks without convolutions are
also competitive \citep{li_constructing_2015,sainath_convolutional_2015,sak_fast_2015}.
Deep ConvNets also contributed to the spectacular success of AlphaGo, an artificial intelligence that beat the world
champion in the game of Go \citep{silver_mastering_2016}.

An attractive property of ConvNets that was leveraged in many previous applications is that they are well suited for 
end-to-end learning, i.e., learning from the raw data without any \emph{a priori} feature selection.
End-to-end learning might be especially attractive in brain-signal decoding, as not all relevant features can be 
assumed to be known a priori.
Hence, in the present study we have investigated how ConvNets of different architectures and designs can be used for
end-to-end learning of EEG recorded in human subjects.

The EEG signal has characteristics that make it different from inputs
that ConvNets have been most successful on, namely images. In contrast
to two-dimensional static images, the EEG signal is a dynamic time
series from electrode measurements obtained on the three-dimensional
scalp surface. Also, the EEG signal has a comparatively low
signal-to-noise ratio, i.e., sources that have no task-relevant
information often affect the EEG signal more strongly than the
task-relevant sources.
These properties could make learning features in
an end-to-end fashion fundamentally more difficult for EEG signals than
for images. Thus, the existing ConvNets architectures from the field of
computer vision need to be adapted for EEG input and the resulting
decoding accuracies rigorously evaluated against more traditional
feature extraction methods. For that purpose, a well-defined baseline is crucial, i.e., a comparison against an 
implementation of a standard EEG decoding method validated on published results for that method.
In light of this, in the present study we addressed two key questions:

\begin{itemize}
\item
  What is the impact of ConvNet \emph{design choices} (e.g., the overall
  network architecture or other design choices such as the type of
  non-linearity used) on the decoding accuracies?
\item
  What is the impact of ConvNet \emph{training strategies} (e.g.,
  training on entire trials or crops within trials) on decoding
  accuracies?
\end{itemize}

To address these questions, we created three ConvNets with different
architectures, with the number of convolutional layers ranging from
2 layers in a ``shallow'' ConvNet over a 5-layer deep ConvNet up to a
31-layer residual network (ResNet).
Additionally, we also created a hybrid ConvNet from the deep and shallow ConvNets.
As described in detail in the
methods section, these architectures were inspired both from existing
``non-ConvNet'' EEG decoding methods, which we embedded in a ConvNet, as
well as from previously published successful ConvNet solutions in the
image processing domain (for example, the ResNet architecture recently
won several image recognition competitions \citep{he_deep_2015}.
All architectures were adapted to the specific
requirements imposed by the analysis of multi-channel EEG data.
To address whether these ConvNets can reach competitive decoding
accuracies, we performed a statistical comparison of their decoding
accuracies to those achieved with decoding based on filter bank common
spatial patterns (FBCSP, \citet{ang_filter_2008,chin_multi-class_2009}), a method that is widely used in EEG decoding 
and has won
several EEG decoding competitions such as BCI Competition IV 2a and 2b.
We analyzed the offline decoding performance on two suitable motor decoding EEG datasets (see 
Section \ref{subsec:datasets} for details).
In all cases, we used only minimal preprocessing to conduct a fair end-to-end comparison of ConvNets and FBCSP.

In addition to the role of the overall network architecture, we
systematically evaluated a range of important design choices. We
focussed on alternatives resulting from recent advances in
machine-learning research on deep ConvNets. Thus, we evaluated potential
performance improvements by using dropout as a novel regularization
strategy \citep{srivastava_dropout:_2014}, intermediate normalization by batch normalization
\citep{ioffe_batch_2015} or exponential linear units as a recently proposed activation function
\citep{clevert_fast_2016}. A comparable analysis of the role of deep ConvNet design
choices in EEG decoding is currently lacking.

In addition to the global architecture and specific design choices which
together define the ``structure'' of ConvNets, another important topic
that we address is how a given ConvNet should be trained on the data.
As with architecture and design, there are several different methodological
options and choices with respect to the training process, such as the
optimization algorithm (e.g., Adam \citep{kingma_adam:_2014}, Adagrad \citep{duchi_adaptive_2011}, etc.),
or the sampling of the training data. Here, we focused on the latter question of sampling the
training data as there is usually, compared to current computer vision
tasks with millions of samples, relatively little data available for EEG
decoding. Therefore, we evaluated two sampling strategies, both
for the deep and shallow ConvNets: training on whole trials or on
multiple crops of the trial, i.e., on windows shifted through the
trials. Using multiple crops holds promise as it increases the amount of
training examples, which has been crucial to the success of deep
ConvNets. Using multiple crops has become standard procedure for
ConvNets for image recognition (see \citep{he_deep_2015,howard_improvements_2013,szegedy_rethinking_2015}, but 
the usefulness of cropped training
has not yet been examined in EEG decoding.

In addition to the problem of achieving good decoding
accuracies, a growing corpus of research tackles the problem of
understanding what ConvNets learn
(see \citet{yeager_effective_2016} for a recent overview). This direction of research may be
especially relevant for neuroscientists interested in using ConvNets ---
insofar as they want to understand what features in the brain signal
discriminate the investigated classes. 
Here we present two novel methods for \emph{feature visualization} that 
we used to gain insights into our ConvNet learned from the neuronal data.

We concentrated on EEG band power features as a target for
visualizations. Based on a large body of literature on
movement-related spectral power modulations
\citep{chatrian_blocking_1959,
pfurtscheller_event-related_1977,
pfurtscheller_occipital_1978,
pfurtscheller_patterns_1989,
pfurtscheller_differentiation_1994,
toro_event-related_1994}, we had clear expectations which band power features should
be discriminative for the different classes; thus our rationale was that
visualizations of these band power features would be particularly useful
to verify that the ConvNets are using actual brain signals. Also, since
FBCSP uses these features too, they allowed us to directly compare
visualizations for both approaches. 
Our first method can be used to
show how much information about a specific feature is retained in the
ConvNet in different layers, however it does not evaluate whether the
feature causally affects the ConvNet outputs. Therefore, we designed our
second method to directly investigate causal effects of the feature
values on the ConvNet outputs. With both visualization methods, it is
possible to derive topographic scalp maps that either show how much
information about the band power in different frequency bands is
retained in the outputs of the trained ConvNet or how much they causally
affect the outputs of the trained ConvNet.

Addressing the questions raised above, in summary the main contributions
of this study are as follows:

\begin{itemize}
\item
  We show for the first time that end-to-end deep ConvNets can reach
  accuracies at least in the same range as FBCSP for decoding
  movement-related information from EEG.
\item
  We evaluate a large number of ConvNet design choices on an EEG
  decoding task, and we show that recently developed methods from the
  field of deep learning such as batch normalization and exponential
  linear units are crucial for reaching high decoding accuracies.
\item
  We show that cropped training can increase the decoding accuracy of
  deep ConvNets and describe a computationally efficient training
  strategy to train ConvNets on a larger number of input crops per EEG
  trial.
\item
  We develop and apply novel visualizations that highly suggest that the
  deep ConvNets learn to use the band power in frequency bands relevant
  for motor decoding (alpha, beta, gamma) with meaningful spatial
  distributions.
\end{itemize}

Thus, in summary, the methods and findings described in this study pave
the way for a widespread application of deep ConvNets for EEG decoding
both in clinical applications and neuroscientific research.
\section{Methods}\label{sec:methods}

We first provide basic definitions with respect to brain-signal decoding as a supervised classification problem used 
in the remaining Methods section.
This is followed by the principles of both filter bank common spatial patterns (FBCSP), the established baseline 
decoding method referred to throughout the present study, and of convolutional neural networks (ConvNets).
Next, we describe the ConvNets developed for this study in detail, including the \emph{design choices} we evaluated.
Afterwards, the training of the ConvNets, including two \emph{training strategies}, are described.
Then we present two novel \emph{visualizations of trained ConvNets} in Section \ref{subsec:visualization}.
Datasets and preprocessing descriptions follow in Section \ref{subsec:datasets}.
Details about statistical evaluation, software and hardware can be found in Supplementary Sections
\ref{subsec:statistics} and \ref{subsec:software-hardware}.

\subsection{Definitions and notation}\label{subsec:definition-notation}

This section more formally defines how brain-signal decoding can be viewed as a supervised classification problem 
and includes the notation used to describe the methods.

\subsubsection{Input and labels}\label{subsec:inputs-labels}

We assume that we are given one EEG data set per subject $i$.
Each dataset is separated into labeled trials (time-segments of the original recording that each belong to one of 
several classes).
Concretely, we are given datasets $D^i=\{(X^1,y^1),...,(X^ {N_i},y^{N_i})\}$ where $N_i$ denotes the total number of 
recorded trials for  
subject $i$.
The input matrix $X^j \in \mathbb{R}^{E\cdot T}$ of trial $j, 1\leq j \leq N_i$ contains the preprocessed signals of 
$E$ 
recorded electrodes 
and $T$ discretized time steps recorded per trial.

The corresponding class label of trial $j$ is denoted by $y^j$.
It takes values from a set of $K$ class labels $L$ that, in our case, correspond to the imagined or executed movements 
performed in each trial, e.g.:
$\forall y^j : y^j \in L = \{l_1 = \text{``Hand (Left)''}, l_2 = \text{``Hand (Right)''}, l_3 = \text{``Feet''}, 
l_4 = \text{``Rest''}\}$.

\subsubsection{Decoder}\label{subsec:decoder}

The decoder $f$ is trained on these existing trials such that it is able to assign the correct label to new unseen 
trials.
Concretely, we aim to train the decoder to assign the label $y^j$ to trial $X^j$  using the output of a parametric 
classifier $f(X^j;\theta): \mathbb{R}^{E\cdot T} \rightarrow L$  with parameters $\theta$.

For the rest of this manuscript we assume that the classifier $f(X^j;\theta)$ is represented by a standard  
machine-learning pipeline decomposed into two parts:
A first part that extracts a (vector-valued) feature representation $\phi(X^j;\theta_\phi)$ with parameters 
$\theta_\phi$ ---  which  could either be set manually (for hand designed features), or learned from the data;
and a second part consisting of a classifier g with parameters $\theta_g$ that is trained using these features, i.e., 
$f(X^j;\theta) = g\big(\phi(X^j;\theta_\phi), \theta_g\big)$.

As described in detail in the following sections, it is important to note that FBCSP and ConvNets differ in how they 
implement this framework:
in short, FBCSP has separated feature extraction and classifier stages, while ConvNets learn both stages jointly.

\subsection{Filter bank common spatial patterns (FBCSP)} \label{subsec:fbcsp}

FBCSP \citep{ang_filter_2008,chin_multi-class_2009} is
a widely-used method to decode oscillatory EEG data, for example, with respect to
movement-related information, i.e., the decoding problem we focus on in
this study. FBCSP was the best-performing method for the BCI competition
IV dataset 2a, which we use in the present study (in the following
called \emph{BCI Competition Dataset},
see Section \ref{subsec:datasets} for a short dataset description). FBCSP also won other
similar EEG decoding competitions \citep{tangermann_review_2012}.
Therefore, we consider FBCSP an adequate benchmark algorithm for the evaluation of the performance of ConvNets in the 
present study.

In the following, we explain the computational steps of FBCSP.
We will refer to these steps when explaining  our shallow ConvNet architecture (see Section 
\ref{subsec:shallow-convnet}), as it is inspired by these steps.

In a supervised manner, FBCSP computes spatial filters (linear combinations of EEG channels) that enhance 
class-discriminative band power features contained in the EEG.
FBCSP extracts and uses these features $\phi(X^j;\theta_\phi)$ (which correspond to the feature representation part in 
Section \ref{subsec:decoder})
to decode the label of a trial in several steps (we will refer back to these steps when we explain the shallow ConvNet):

\begin{enumerate}
\item
  \textbf{Bandpass filtering:} Different bandpass filters are applied to separate the raw EEG signal into different 
  frequency bands.
\item
  \textbf{Epoching:} The continuous EEG signal is cut into trials as explained in Section \ref{subsec:inputs-labels}.
\item
  \textbf{CSP computation:} Per frequency band, the common spatial patterns (CSP) algorithm is applied to extract 
spatial filters.
  CSP aims to extract spatial filters that make the trials discriminable by the power of the spatially filtered trial 
signal (see \citet{koles_spatial_1990,ramoser_optimal_2000,blankertz_optimizing_2008} for more 
details on the computation).
  The spatial filters correspond to the learned parameters $\theta_\phi$ in FBCSP.
\item
  \textbf{Spatial filtering:} The spatial filters computed in Step 2 are applied to the EEG signal.
\item
  \textbf{Feature construction:} Feature vectors $\phi(X^j;\theta_\phi)$ are constructed from the filtered signals:
  Specifically, feature vectors are the log-variance of the spatially filtered trial signal for each frequency band and 
for each spatial filter.
\item
  \textbf{Classification:} A classifier is trained to predict per-trial labels based on the feature vectors.
\end{enumerate}

For details on our FBCSP implementation, see Supplementary Section \ref{subsec:supp-fbcsp}.

\subsection{Convolutional neural networks}\label{subsec:convnets}

In the following sections, we first explain the basic ideas of ConvNets.
We then describe architectural choices for ConvNets on   EEG, including how to represent the EEG input for a ConvNet, the three different ConvNet architectures used in this study and several specific design choices that we evaluated for these architectures.
Finally, we describe how to train the ConvNets, including the description of a trial-wise and a cropped training 
strategy for our EEG data.

\subsubsection{Basics}\label{subsec:convnet-basics}

Generally, ConvNets combine two ideas useful for many learning tasks on natural signals, such as images and audio  
signals.
These signals often have an inherent hierarchical structure (e.g., images typically consist of edges that together
form simple shapes which again form larger, more complex shapes and so on).
ConvNets can learn local non-linear features (through convolutions and nonlinearities) and represent 
higher-level features as compositions of lower level features (through multiple layers of processing).
In addition, many ConvNets use pooling layers which create a coarser intermediate feature representation and can make 
the ConvNet more translation-invariant.
For further details see 
\citet{lecun_deep_2015,goodfellow_deep_2016,schmidhuber_deep_2015}.

\subsection{ConvNet architectures and design choices}\label{subsec:convnet-architectures}

\subsubsection{Input representation}\label{subsec:convnet-input-representation}

The first important decision for applying ConvNets to EEG decoding is how to represent the input $X^j \in 
\mathbb{R}^{E\cdot T}$.
One possibility would be to represent the EEG as a time series of topographically organized images, i.e., of the 
voltage distributions across the (flattened) scalp surface (this has been done for ConvNets that get power spectra as 
input \citep{bashivan_learning_2016}.
However, EEG signals are assumed to approximate a linear superposition of spatially global voltage patterns caused by 
multiple dipolar current sources in the brain \citep{nunez_electric_2006}.
Unmixing of these global patterns using a number of spatial filters is therefore typically applied to the whole set of 
relevant electrodes as a basic step in 
many successful examples of EEG decoding
\citep{ang_filter_2008,blankertz_optimizing_2008,rivet_xdawn_2009}.

In this view, all relevant EEG modulations are global in nature, due to the physical origin of the non-invasive EEG and 
hence there would be no obvious
hierarchical compositionality of local and global EEG modulations \emph{in space}.
In contrast, there is an abundance of evidence that the EEG is organized across multiple time
scales, such as in nested oscillations 
\citep{canolty_high_2006,monto_very_2008,schack_phase-coupling_2002,vanhatalo_infraslow_2004}
involving both local and global modulations \emph{in time}.
In light of this, we designed ConvNets in a way that they can learn spatially global unmixing filters in the entrance 
layers, as well as temporal hierarchies
of local and global modulations in the deeper architectures.
To this end we represent the input as a 2D-array with the number of time steps as the width and the number of 
electrodes as the height.
This approach also significantly reduced the input dimensionality compared with the ``EEG-as-an-image'' approach.

\subsubsection{Deep ConvNet for raw EEG signals}\label{subsec:deep-convnet}

\begin{figure}
  \begin{center}
  \includegraphics[max size={\linewidth}{0.61\paperheight}]{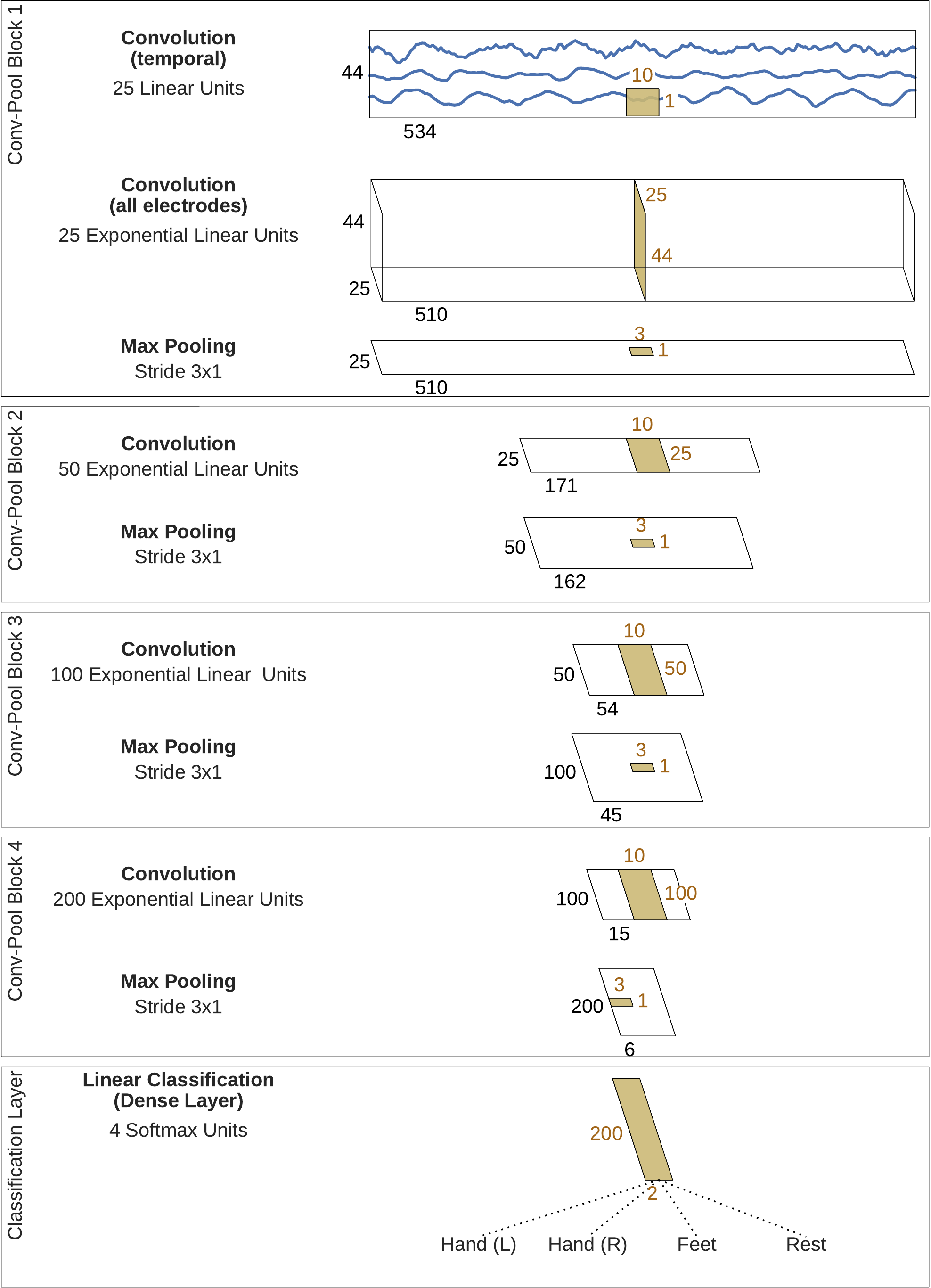}
  \end{center}
    \caption{\textbf{Deep ConvNet architecture.} 
EEG input (at the top) is progressively transformed towards the bottom,
until the final classifier output. Black cuboids: inputs/feature maps;
brown cuboids: convolution/pooling kernels. The corresponding sizes are
indicated in black and brown, respectively. Note that in
this schematics, proportions of maps and kernels are only approximate.}
    \label{fig:deep-net}
\end{figure}
To tackle the task of EEG decoding we designed a deep ConvNet architecture inspired by successful architectures in 
computer vision, as for example described in \citet{krizhevsky_imagenet_2012}.
The requirements for this architecture are as follows:
We want a model that is able to extract a wide range of features and is not restricted to specific feature types
\citep{hertel_deep_2015}.
We were interested in such a generic architecture for two reasons:
1) we aimed to uncover whether such a generic ConvNet designed only with minor expert knowledge can reach competitive 
accuracies, and,
2) to lend support to the idea that standard ConvNets can be used as a general-purpose tool for brain-signal decoding 
tasks.
As an aside, keeping the architecture generic also increases the chances that ConvNets for brain decoding 
can directly profit from future methodological advances in deep learning.

Our deep ConvNet had four convolution-max-pooling blocks, with a special first block designed to handle EEG input (see 
below),
followed by three standard convolution-max-pooling blocks and a dense softmax classification layer 
(see Figure \ref{fig:deep-net}).
The first convolutional block was split into two convolutional layers in order to better handle the large number of 
input channels --- one input channel per electrode compared to three input channels (one per color) in rgb-images.
The convolution was split into a first convolution across time and a second convolution across space (electrodes);
each filter in these steps has weights for all electrodes (like a CSP spatial filter) and for the filters of the 
preceding temporal convolution (like any standard intermediate convolutional layer).
Since there is no activation function in between the two convolutions, they could in principle be combined into one 
layer.
Using two layers however implicitly regularizes the overall convolution by forcing a separation
of the linear transformation into a combination of two (temporal and spatial) convolutions.
This splitted convolution was evaluated against a single-step convolution in our experiments 
(see Section \ref{subsec:convnet-design-choices}).

We used exponential linear units (ELUs, $f(x)=x$ for $x > 0$ and $f(x) = e^x-1$ for $x <= 0$
\citep{clevert_fast_2016}) as activation functions (we also evaluated Rectified Linear Units (ReLUs, $f(x) = 
max(x,0)$), as a less 
recently proposed alternative, see Section \ref{subsec:convnet-design-choices}).

\subsubsection{Shallow ConvNet for raw EEG signals}\label{subsec:shallow-convnet}

\begin{figure}
  \begin{center}
  \includegraphics[max size={\linewidth}{0.3\paperheight}]{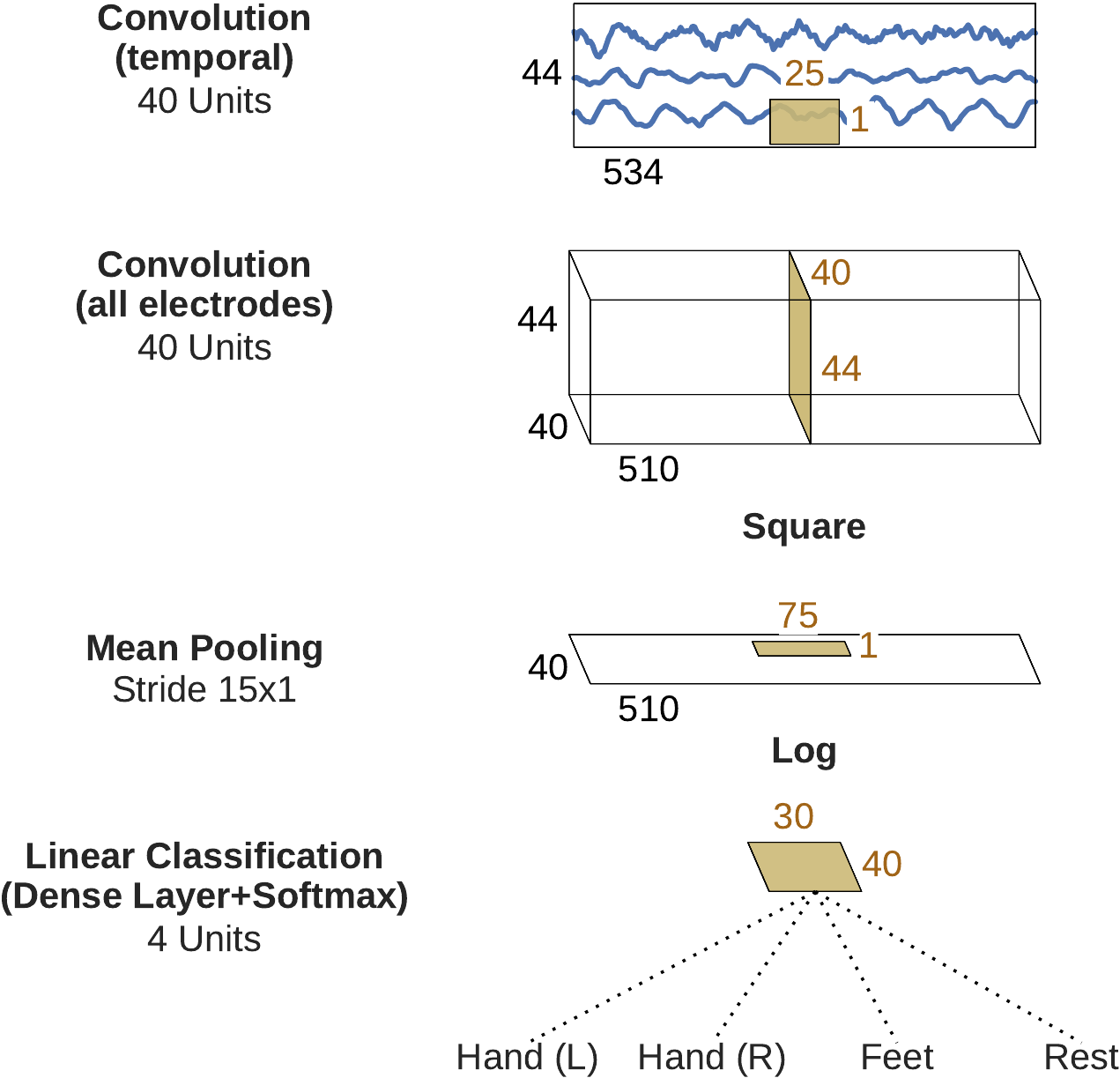}
  \end{center}
    \caption{\textbf{Shallow ConvNet architecture.} 
    Conventions as in Figure \ref{fig:deep-net}.}
    \label{fig:shallow-net}
\end{figure}
We also designed a more shallow architecture referred to as shallow ConvNet, inspired by the FBCSP pipeline (see Figure 
\ref{fig:shallow-net}),
specifically tailored to decode band power features.
The transformations performed by the shallow ConvNet are similar to the transformations of FBCSP (see Section 
\ref{subsec:fbcsp}).
Concretely, the first two layers of the shallow ConvNet perform a temporal and a spatial convolution, as in the deep 
ConvNet.
These steps are analogous to the bandpass and CSP spatial filter steps in FBCSP.
In contrast to the deep ConvNet, the temporal convolution of the shallow ConvNet had a larger kernel size (25 vs 10),
allowing a larger range of transformations in this layer (smaller kernel sizes for the shallow ConvNet led to 
lower accuracies in preliminary experiments).
After the two convolutions of the shallow ConvNet, a squaring nonlinearity, a mean pooling layer and a 
logarithmic activation function followed;
together these steps are analogous to the trial log-variance computation in FBCSP (we note that these steps were 
not used in the deep ConvNet).
In contrast to FBCSP, the shallow ConvNet embeds all the computational steps in a single network, and thus all steps 
can 
be optimized jointly (see Section \ref{subsec:convnet-training}).
Also, due to having several pooling regions within one trial, the shallow ConvNet can learn a temporal structure of the 
band power changes within the trial, which was shown to help classification in prior work 
\citep{sakhavi_parallel_2015}.

\subsubsection{Design choices for deep and shallow
ConvNet}\label{subsec:convnet-design-choices}

For both architectures described above we evaluated several design choices.
We evaluated architectural choices which we expect to have a potentially large impact on the decoding accuracies and/or 
from which we hoped to gain insights into the behavior of the ConvNets.
Thus, for the deep ConvNet, we compared the design aspects listed in Table \ref{table:design-choices}.

\setlength\dashlinedash{0.6pt}
\setlength\dashlinegap{1.5pt}
\setlength\arrayrulewidth{0.6pt}
\begin{table}[H]
\centering\footnotesize
\begin{tabular}{p{0.15\linewidth}p{0.15\linewidth}p{0.15\linewidth}p{0.45\linewidth}}\toprule
\textbf{\normalsize Design aspect} & \textbf{\normalsize Our choice} & \textbf{\normalsize Variants} & 
\textbf{\normalsize Motivation} \\\midrule
 Activation  ~~~~~ functions & ELU & square, ReLU & 
\multirow{2}{\linewidth}{We expected these choices to be sensitive to the type of feature (e.g.,
signal phase or power), since squaring and mean pooling results in mean
power (given a zero-mean signal). Different features may play different
roles in the low-frequency components vs. the higher frequencies (see Section \ref{subsec:datasets}).}\\
\cdashline{1-3}
Pooling mode & max & mean & \\ \ \\ \ \\ \ \\ \ \\
\hdashline
Regularization and intermediate normalization & Dropout + batch normalization + a new tied loss function (explanations
see text) & Only batch normalization, only dropout, neither of both, no tied loss & 
We wanted to investigate whether recent deep learning advances improve
accuracies and check how much regularization is required.\\
\hdashline
Factorized temporal convolutions & One 10x1 convolution per convolutional layer & Two 6x1 convolutions per 
convolutional 
layer & Factorized convolutions are used by other successful ConvNets (see \citet{szegedy_rethinking_2015})\\
\hdashline
Splitted vs one-step convolution & Splitted convolution in first layer (see Section \ref{subsec:deep-convnet}) & 
one-step  convolution in first layer &
Factorizing convolution into spatial
and temporal parts may improve accuracies for the large number of EEG input
channels (compared with three rgb color channels of regular image datasets).\\ \bottomrule
\hline
\end{tabular}
\caption{\textbf{Evaluated design choices.} Design choices we evaluated for our
convolutional networks. ``Our choice'' are the choices we used when
evaluating ConvNets in the remainder of this manuscript, e.g., vs FBCSP. Note that these design choices have not been
evaluated in prior work, see Supplementary Section \ref{supp:subsec-related-work} }
\label{table:design-choices}
\end{table}

In the following, we give additional details for some of these aspects.
Batch normalization standardizes intermediate 
outputs of the network to zero mean and unit 
variance for a batch of training examples \citep{ioffe_batch_2015}.
This is meant to facilitate the optimization by keeping the inputs of layers closer to a normal distribution during 
training.
We applied batch normalization, as recommended in the original paper 
\citep{ioffe_batch_2015}, to the output of convolutional layers before the nonlinearity.
Dropout randomly sets some inputs for a layer to zero in each training update.
It is meant to prevent co-adaption of different units and can be seen as analogous to training an ensemble of networks.
We drop out the inputs to all convolutional layers after the first with a probability of 0.5.
Finally, our new tied loss function is designed to further regularize our cropped training (see Section 
\ref{subsec:cropped-training} for an explanation).

We evaluated the same design aspects for the shallow ConvNet, except for the following differences:

\begin{itemize}
\item
  The baseline methods for the activation function and pooling mode were chosen as ``squaring nonlinearity'' and ``mean 
pooling'', motivation is given in Section \ref{subsec:shallow-convnet}.
\item
  We did not include factorized temporal convolutions into the  comparison, as the longer kernel lengths of the shallow 
ConvNet make these convolutions less similar to other successful ConvNets anyways.
\item
  We additionally compared the logarithmic nonlinearity after the pooling layer with a square root nonlinearity to 
check 
if the
  logarithmic activation inspired by FBCSP is better than traditional L2-pooling.
\end{itemize}

\subsubsection{Hybrid ConvNet}\label{subsec:hybrid-convnet}

Besides the individual design choices for the deep and shallow ConvNet, a natural question to ask is whether both 
ConvNets can be combined into a single ConvNet.
Such a hybrid ConvNet could profit from the more specific feature extraction of the shallow ConvNet as well as from the 
more generic feature extraction of the deep ConvNet.
Therefore, we also created a hybrid ConvNet by fusing both networks after the final layer.
Concretely, we replaced the four-filter softmax classification layers of both ConvNets by 60- and 40-filter ELU layers 
for the deep and shallow ConvNet respectively.
The resulting 100 feature maps were concatenated and used as the input to a new softmax classification layer.
We retrained the whole hybrid ConvNet from scratch and did not use any pretrained deep or shallow ConvNet parameters.

\subsubsection{Residual ConvNet for raw EEG signals}\label{subsec:resnet}

\begin{figure}
\centering
\includegraphics{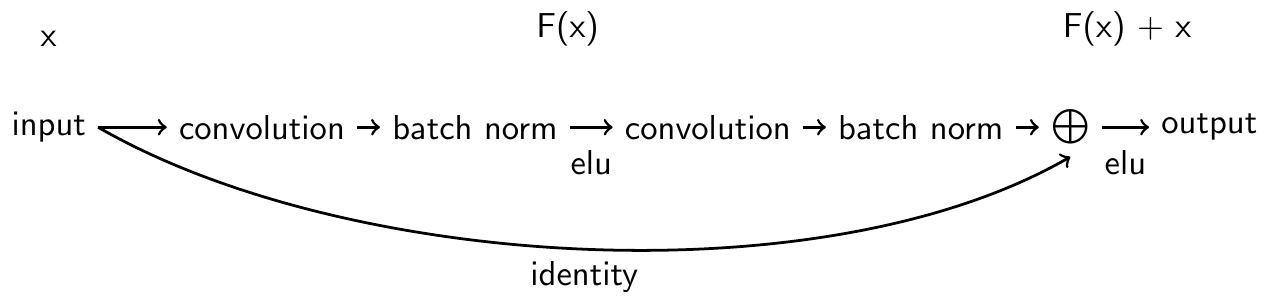}
\caption{\textbf{Residual Block}. Residual block used in the ResNet architecture and as described in
original 
paper \citep{he_deep_2015}, see Figure 2) with identity shortcut option A, except
using ELU instead of ReLU nonlinearities.
See Section \ref{subsec:resnet} for explanation.}
\label{fig:residual-block}
\end{figure}

In addition to the shallow and deep ConvNets, we evaluated another network architecture:
Residual networks (ResNets), a ConvNet architecture that recently won several benchmarks in the computer vision field
\citep{he_deep_2015}.
ResNets typically have a very large number of layers and we wanted to investigate whether similar networks with 
more layers also result in good performance in EEG decoding.
ResNets add the input of a convolutional layer to the output of the same layer, to the effect that the convolutional 
layer only has to learn to output a residual that changes the previous layers output (hence the name residual network).
This allows ResNets to be successfully trained with a much larger number of layers than traditional convolutional 
networks \citep{he_deep_2015}.
Our residual blocks are the same as described in the original paper (see Figure \ref{fig:residual-block}).

Our ResNet used exponential linear unit activation functions \citep{clevert_fast_2016} throughout the network (same as 
the deep ConvNet) and also starts with a splitted temporal and spatial convolution (same as the 
deep and shallow ConvNets), followed by 14 residual blocks, mean pooling and a
final softmax dense classification layer (for further details, see Supplementary Section 
\ref{subsec:supp-resnet-architecture}).

\subsection{ConvNet training}\label{subsec:convnet-training}

In this section, we first give a definition of how ConvNets are trained in general.
Second, we describe two ways of extracting training inputs and training labels from the EEG data, which result in a 
trialwise and a
cropped training strategy.

\subsubsection{Definition}

To train a ConvNet, all parameters (all weights and biases) of the ConvNet are trained jointly.
Formally, in our supervised classification setting, the ConvNet computes a function from input data to one real number 
per class,
$f(X^j;\theta): \mathbb{R}^{E\cdot T}\rightarrow\mathbb{R}^K$, where $\theta$ are the parameters of the function,
$E$ the number of electrodes, $T$ the number of timesteps and $K$ the number of possible output labels.
To use ConvNets for classification, the output is typically transformed to conditional probabilities of a label $l_k$ 
given the
input $X^j$ using the softmax function:
$p(l_k|f(X^j;\theta))= \frac{exp(f_k(X^j;\theta))}{\sum_{k=1}^{K}{exp(f_k(X^j;\theta))}}$.
In our case, since we train per
subject, the softmax output gives us a subject-specific conditional
distribution over the $K$ classes. Now we can train the entire ConvNet to assign high
probabilities to the correct labels by minimizing the sum of the
per-example losses:
\begin{align}
\theta^*=\arg \min_\theta \sum_{j=1}^N loss\Big(y^j, p\big(l_k|f_k(X^j;\theta)\big)\Big)
\end{align}, where 
\begin{align}
loss\Big(y^j, p\big(l_k|f_k(X^j;\theta)\big)\Big) = \sum_{k=1}^{K}{-log\Big(p\big(l_k|f_k(X^j;\theta)\big)\Big) \cdot 
\delta(y^j=l_k)} 
\label{eq:untied-loss}
\end{align}
is the negative log likelihood of the labels.
As is common for training ConvNets, the parameters are optimized via mini-batch stochastic gradient descent using 
analytical gradients computed via backpropagation (see
\citet{lecun_deep_2015} for an explanation in the context of ConvNets and Section \ref{subsec:optimization} in this 
manuscript for details on the optimizer used in this study).

This ConvNet training description is connected to our general EEG decoding definitions from Section 
\ref{subsec:definition-notation} as follows. The function that the ConvNet 
computes can be viewed as consisting of a feature extraction function and a classifier 
function:
The feature extraction function $\phi(X^j;\theta_\phi)$ with parameters $\theta_\phi$ is computed by all layers up to
the penultimate layer.
The classification function $g\big(\phi(X^j;\theta_\phi), \theta_g\big)$ with parameters $\theta_g$, which uses the 
output of the feature extraction function as input, is computed by the final classification layer.
In this view, one key advantage of ConvNets becomes clear:
With the joint optimization of both functions, a ConvNet learns both, a descriptive feature representation for the task 
as well as a discriminative classifier.
This is especially useful with large datasets, where it is more likely that the ConvNet learns to extract useful
features and does not just overfit to noise patterns.
For EEG data, learning features can be especially valuable since there may be unknown discriminative features or at 
least discriminative features that are not used by more traditional feature extraction methods such as FBCSP.

\subsubsection{Input and labels}

In this study, we evaluated two ways of defining the input examples and target labels that the ConvNet is trained on.
First, a trial-wise strategy that uses whole trials as input and per-trial labels as targets.
Second, a cropped training strategy that uses crops, i.e., sliding time windows within the trial as input and 
per-crop labels  as targets (where the label of a crop is identical to the label of the trial the crop was extracted 
from).

\subsubsection{Trial-wise training}

The standard trial-wise training strategy uses the whole duration of the trial and is therefore similar to how FBCSP is 
trained.
For each trial, the trial signal is used as input and the corresponding trial label as target to train the ConvNet.
In our study, for both datasets we had 4.5-second trials (from 500 ms before trial start cue until trial 
end cue, as that worked best in preliminary experiments) as the input to the network. This led to 288 training 
examples per subject for the BCI Competition Dataset and about 880 training examples per subject on the High-Gamma 
Dataset 
after their respective train-test split.

\subsubsection{Cropped training}\label{subsec:cropped-training}

\begin{figure}
\begin{subfigure}[t]{\linewidth}
\centering
\includegraphics[max size={\linewidth}{0.2\paperheight}]{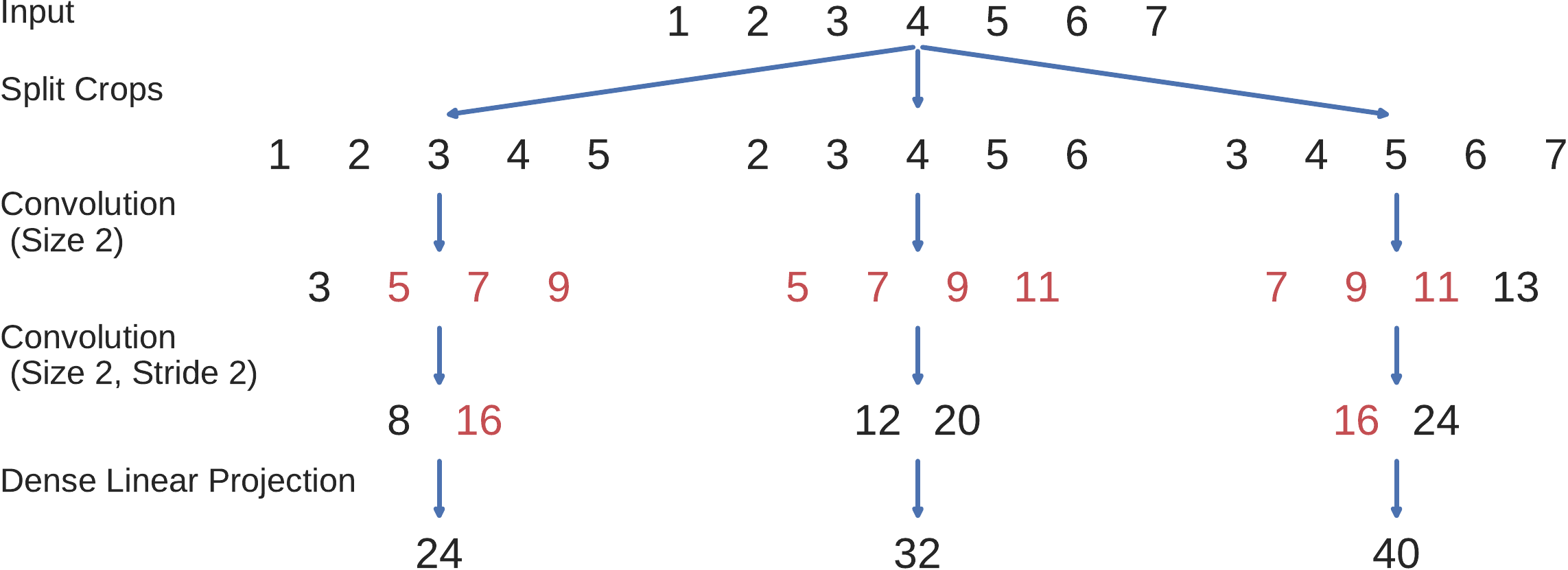}
\caption{\footnotesize{Naïve implementation by first splitting the trial into crops and
  passing the crops through the ConvNet independently.}}
\label{fig:cropped-naive}
\end{subfigure}%
\vspace{0.5cm}
\\
\begin{subfigure}[t]{\linewidth}
\centering
\includegraphics[max size={\linewidth}{0.28\paperheight}]{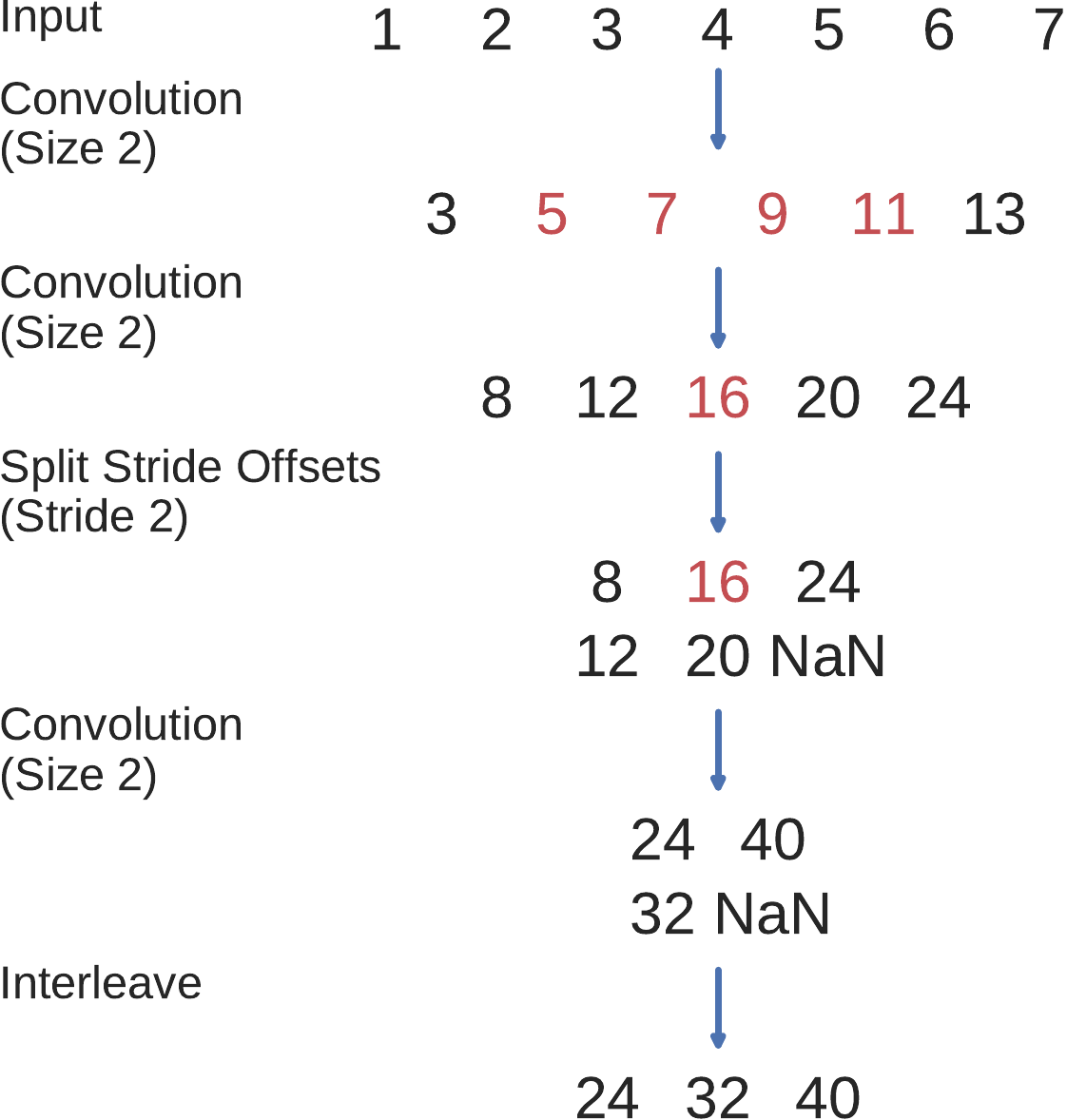}
\caption{\footnotesize{Optimized implementation, computing the outputs for each crop in a
single forward pass. Strides in the original ConvNet are handled by
separating intermediate results that correspond to different stride
offsets, see the split stride offsets step. NaNs are only needed to pad
all intermediate outputs to the same size and are removed in the end.
The split stride step can simply be repeated in case of further layers
with stride. We interleave the outputs only after the final
predictions, also in the case of ConvNets with more layers.}}
\label{fig:cropped-smarter}
\end{subfigure}
\caption{\textbf{Multiple-crop prediction used for cropped training.}
In this toy example, a trial with the sample values 1,2,3,4,5,6,7 is cut into three crops of length 5 and these crops 
are passed through a convolutional network with two convolutional layers and one dense layer.
The convolutional layers both have kernel size 2, the second one additionally uses a stride of 2.
Filters for both layers and the final dense layer have values 1,1.
Red indicates intermediate outputs that were computed multiple times in the naïve implementation.
Note that both implementations result in the same final outputs.}
\label{fig:cropped-explanation}
\end{figure}

The cropped training strategy uses crops, i.e., sliding input windows within the trial,
which leads to many more training examples for the network than the trial-wise training strategy.
We adapted this strategy from convolutional neural networks for object recognition in images,
where using multiple crops of the input image is a standard procedure to increase decoding accuracy (see for example
\citet{he_deep_2015} and \citet{szegedy_rethinking_2015}).

In our study, we used crops of about 2 seconds as the input. We adopt a cropping approach, which leads to the 
largest possible number of crops by creating one crop per sample (by sample, we mean a timestep in our EEG trial time 
series).
More formally, given an original trial $X^j \in \mathbb{R}^{E\cdot T}$ with $E$ electrodes and $T$ timesteps,
we create a set of crops with crop size $T'$ as timeslices of the trial: 
$C^j=\{X^j_{1..E,t..t+T}|t\in 1..T-T'\}$. 
All of these $T-T'$ crops are new training data examples for our decoder and
will get the same label $y^j$ as the original trial.

This aggressive cropping has the aim to force the ConvNet into using features that are present in all crops of the 
trial,
since the ConvNet can no longer use the differences between crops and the global temporal structure of the features in
the complete trial.
We collected crops starting from 0.5 seconds before trial start (first crop from 0.5 seconds before to 1.5 seconds 
after trial start), with the last crop ending 4 seconds after the trial start
(which coincides with the trial end, so the last crop starts 2 seconds before the trial and continues to the trial end).
Overall, this resulted in 625 crops and therefore 625 label predictions per trial.
The mean of these 625 predictions is used as the final prediction for the trial during the test phase.
During training, we compute a loss for each prediction.
Therefore, cropped training increases our training set size by a factor of 625, albeit with highly correlated training 
examples.
Since our crops are smaller than the trials, the ConvNet input size is also smaller (from about 1000 input samples to
about 500 input samples for the 250 Hz sampling rate), while all other hyperparameters stay the same.

To reduce the computational load from the increased training set size,
we decoded a group of neighboring crops together and reused intermediate convolution outputs.
This idea has been used in the same way to speed up ConvNets that make predictions for each pixel in an image
\citep{giusti_fast_2013,nasse_face_2009,sermanet_overfeat:_2014,shelhamer_fully_2016}.
In a nutshell, this method works by providing the ConvNet with an input that contains several crops and computing the
predictions for all crops in a single forward pass (see Figure \ref{fig:cropped-explanation} for 
an explanation).
This cropped training method leads to a new hyperparameter: the number of crops that are processed at the same time.
The larger this number of crops, the larger the speedup one can get (upper bounded by the size of one crop, see
\citet{giusti_fast_2013} for a more detailed speedup analysis on images), at the cost of increased memory consumption.
A larger number of crops that are processed at the same time during training also implies parameter updates from 
gradients computed on a larger number of crops from the same trial during mini-batch stochastic gradient descent, with 
the risk of less stable training.
However, we did not observe substantial accuracy decreases when enlarging the number of simultaneously processed crops 
(this stability was also observed for images \citep{shelhamer_fully_2016})
and in the final implementation we processed about 500 crops in one pass, which corresponds to passing the ConvNet an 
input of about 1000 samples, twice the 500 samples of one crop.
Note that this method only results in exactly the same predictions as the naïve method when using valid convolutions 
(i.e., no padding).
For padded convolutions (which we use in the residual network described in Section 
\ref{subsec:resnet}), the method no longer results in the same predictions, so it cannot be used to speed up 
predictions for individual samples anymore.
However, it can still be used if one is only interested in the average prediction for a trial as we are in this study.

To further regularize ConvNets trained with cropped training, we designed a new objective function, which penalizes
discrepancies between predictions of neighboring crops.
In this \textit{tied sample loss function}, we added the cross-entropy of two neighboring predictions to the usual loss 
of of negative log likelihood of the labels. 
So, denoting the prediction $p\big(l_k|f_k(X^j_{t..t+T'};\theta)\big)$ for crop 
$X^j_{t..t+T'}$ from time step $t$ to $t+T'$ by $p_{f,k}(X^j_{t..t+T'})$, the loss now also depends on the
prediction for the next crop  $p_{f,k}(X^j_{t..t+T'+1})$ and changes from equation \ref{eq:untied-loss} to:
 
 \begin{equation}\label{eq:tied-loss}
\begin{split}
loss\big(y^j, p_{f,k}(X^j_{t..t+T'})\big)& =\sum_{k=1}^{K}-log\big(p_{f,k}(X^j_{t..t+T'})\big) \cdot \delta(y^j=l_k) 
\quad +\\
& \quad \sum_{k=1}^{K}-log\big(p_{f,k}(X^j_{t..t+T'})\big) \cdot p_{f,k}(X^j_{t..t+T'+1})
\end{split}
\end{equation}

This is meant to make the ConvNet focus on features which are stable for several neighboring input crops.

\subsubsection{Optimization and early stopping}\label{subsec:optimization}

As optimization method, we used Adam \citep{kingma_adam:_2014} together with a specific early stopping method,
since this consistently yielded good accuracies in our experiments.
For details on Adam and our early stopping strategy, see Supplementary Section \ref{subsec:supp-optimization}.

\subsection{Visualization}\label{subsec:visualization}

\subsubsection{Correlating Input Features and Unit Outputs: Network
Correlation Maps}\label{subsec:ncm-visualizations}

\begin{figure}
  \begin{center}
  \includegraphics[max size={\linewidth}{0.2\paperheight}]{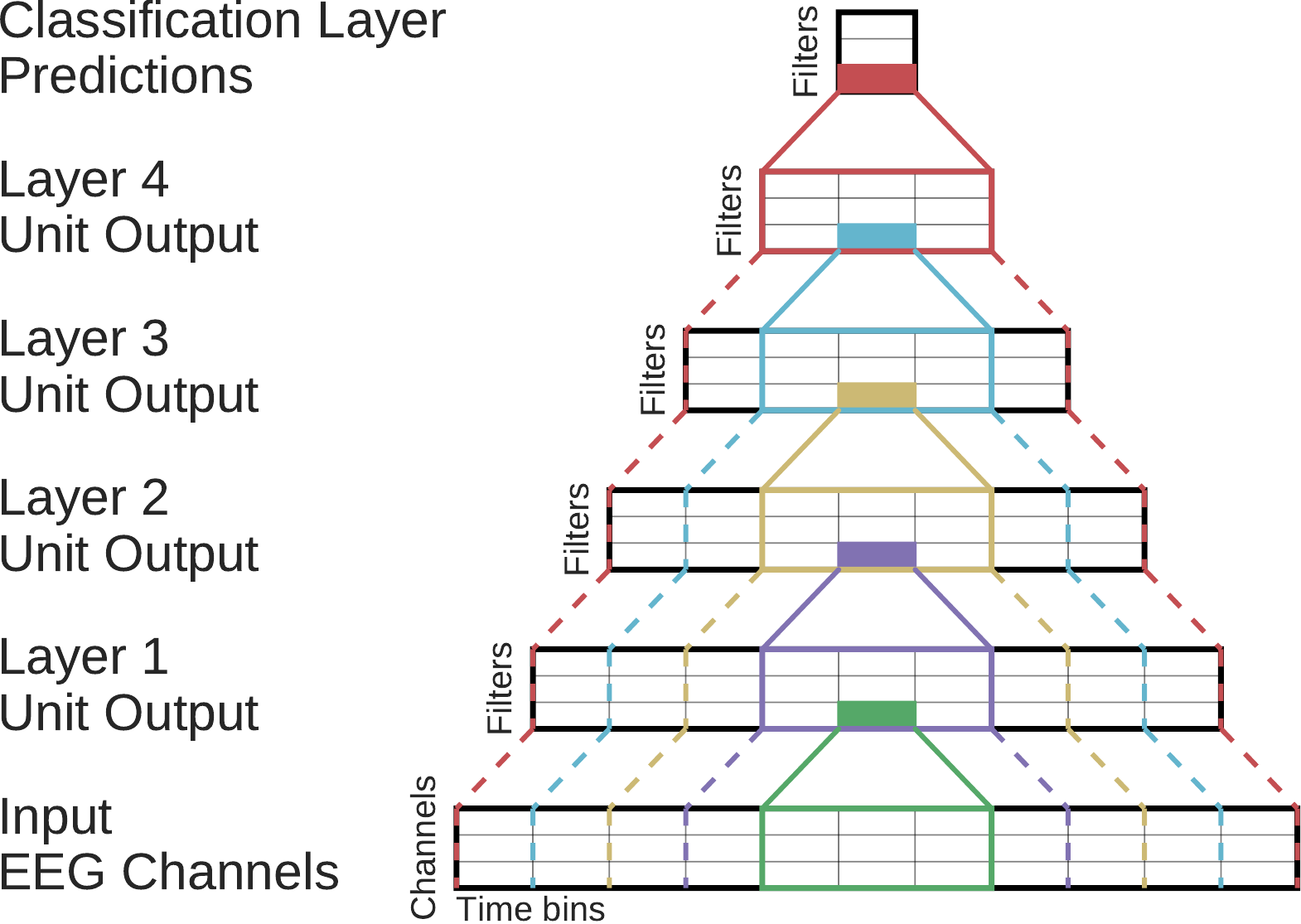}
  \end{center}
    \caption{\textbf{ConvNet Receptive Fields Schema}. 
    Showing the outputs, inputs and receptive fields of one unit per layer. 
    Colors indicate different units.
    Filled rectangles are individual units, solid lines indicate their direct input from the layer before.
    Dashed lines indicate the corresponding receptive field in all previous layers including the original input layer.
    The receptive field of a unit contains all inputs that are used to compute the unit's output.
    The receptive fields get larger with increasing depth of the layer.
    Note that this is only a schema and exact dimensions are not meaningful in this figure.}
    \label{fig:receptive-fields}
\end{figure}

As described in the Introduction, currently there is a great interest in understanding how ConvNets learn to solve 
different tasks.
To this end, methods to visualize functional aspects of ConvNets can be helpful and the development of such methods is
an active area of research.
Here, we wanted to delineate what brain-signal features the ConvNets used and in which layers they extracted these 
features.
The most obvious restriction on possible features is that units in individual layers of the ConvNet can only extract 
features
from samples that they have ``seen'', i.e., from their so-called \emph{receptive field} (see Figure 
\ref{fig:receptive-fields}).
A way to further narrow down the possibly used features is to use domain-specific prior knowledge and to
investigate whether known class-discriminative features are learned by the ConvNet.
Then it is possible to compute a feature value for all receptive fields of all individual units for each of these
class-discriminative features and to measure how much this feature affects the unit output, for example by computing 
the correlation between feature values and unit outputs.

In this spirit, we propose input-feature unit-output correlation maps as
a method to visualize how networks learn spectral amplitude features.
It is known that the amplitudes, for example of the alpha, beta and gamma
bands, provide class-discriminative information for motor tasks 
\citep{ball_movement_2008,pfurtscheller_central_1981,pfurtscheller_evaluation_1979}.
Therefore, we used the mean envelope values for several frequency bands as feature values.
We correlated these values inside a receptive field of a unit, as a measure of its total spectral amplitude,
with the corresponding unit outputs to gain insight into how much these amplitude features are used by the ConvNet.
Positive or negative correlations that systematically deviate from those found in an untrained net imply that 
the ConvNet learned to create representations that contain more information about these 
features than before training.

A limitation of this approach is that it does not distinguish between correlation and causation
(i.e., whether the change in envelope caused the change in the unit output, or whether another
feature, itself correlated to the unit output, caused the change).
Therefore, we propose a second visualization method, where we perturbed the amplitude of existing inputs 
and observed the change in predictions of the ConvNets. 
This complements the first visualization and we refer to this method as input-perturbation network-prediction 
correlation map.
By using artificial perturbations of the data, they provide insights in whether changes in specific feature amplitudes
cause the network to  change its outputs.
For details on the computation of both NCM methods and a ConvNet-independent visualization, see Supplementary Section 
\ref{subsec:supp-visualization-methods}.

\subsection{Data sets and preprocessing}\label{subsec:datasets}
We evaluated decoding accuracies on two EEG datasets, a smaller public dataset (BCI Competition IV dataset 2a)
for comparing to previously published accuracies and a larger new dataset acquired in our lab for
evaluating the decoding methods with a larger number of training trials (approx. 880 trials per subject, 
compared to 288 trials in the public set). For details on the datasets, see Supplementary Section 
\ref{subsec:supp-datasets}.

\subsubsection{EEG preprocessing and evaluating different frequency
bands}\label{subsec:preprocessing}

We only minimally preprocessed the datasets to allow the ConvNets to
learn any further transformations themselves.
In addition to the full-bandwidth (0--$f_{end}$-Hz) dataset, we analyzed 
data high-pass filtered above 4 Hz (which we call 4--$f_{end}$-Hz dataset).
Filtering was done with a causal 3rd order Butterworth filter.
We included the 4--$f_{end}$-Hz dataset since the highpass filter should make it less probable
that either the networks or FBCSP would use class-discriminative eye movement artifacts to decode the behavior classes,
as eye movements generate most power in such low frequencies \citep{gratton_dealing_1998}.
We included this analysis as for the BCI Competition Dataset, special care to avoid decoding eye-related signals was 
requested from the publishers of the dataset \citep{brunner_bci_2008}.
For details on other preprocessing steps, see Supplementary Section \ref{subsec:supp-preprocessing}.

\section{Results}\label{sec:results}
\subsection{Validation of FBCSP baseline}
\begin{result}
FBCSP baseline reached same results as previously reported in the literature \label{subsec:results-baseline}
\end{result}

As a first step before moving to the evaluation of ConvNet decoding, we
validated our FBCSP implementation, as this was the baseline we compared the ConvNets results against.
To validate our FBCSP implementation, we compared its accuracies to those published in the literature for the
BCI competition IV dataset 2a (called BCI Competition Dataset in the following) 
\citep{sakhavi_parallel_2015}.
Using the same 0.5--2.5 s (relative to trial onset) time window, we reached an accuracy of 67.6\%, statistically not
significantly different from theirs (67.0\%, p=0.73, Wilcoxon signed-rank test).
Note however, that we used the full trial window for later experiments with convolutional networks, i.e., from 0.5--4 
seconds.
This yielded a slightly better accuracy of 67.8\%, which was still not statistically significantly different from the 
original results
on the 0.5--2.5 s window (p=0.73).
For all later comparisons, we use the 0.5--4 seconds time window on all datasets.

\subsection{Architectures and design choices}

\begin{result}
ConvNets reached FBCSP accuracies \label{subsec:results-fbcsp-convnets}
\end{result}

\begin{figure}
\centering
\includegraphics[max size={1\linewidth}{0.56\paperheight}]{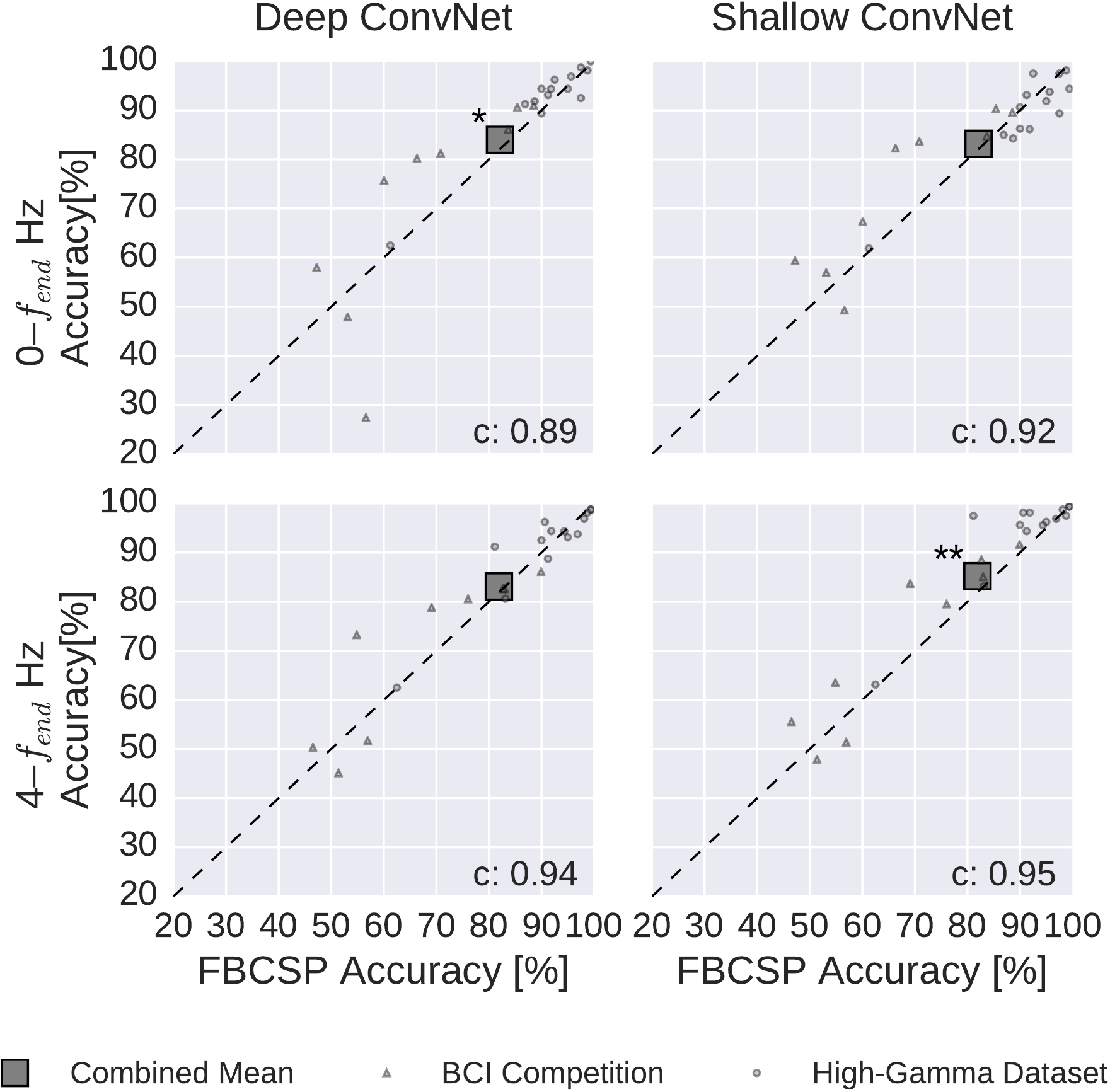}
\caption{\textbf{FBCSP vs. ConvNet decoding accuracies.} Each small marker represents
accuracy of one subject, the large square markers represent average
accuracies across all subjects of both datasets. Markers above the
dashed line indicate experiments where ConvNets performed better than
FBCSP and opposite for markers below the dashed line. Stars indicate
statistically significant differences between FBCSP and ConvNets
(Wilcoxon signed-rank test, p\textless{}0.05: *, p\textless{}0.01: **,
p\textless{}0.001=***). Bottom left of every plot: linear correlation
coefficient between FBCSP and ConvNet decoding accuracies. Mean
accuracies were very similar for ConvNets and FBCSP, the (small)
statistically significant differences were in direction of the ConvNets.}
\label{fig:results-fbcsp-convnets}
\end{figure}

\begin{table}
\centering 
\begin{tabular}{lllllll}\toprule
\textbf{\normalsize Dataset} & 
\parbox{0.15\linewidth}{\textbf{\normalsize Frequency\\range [Hz]}} & 
\textbf{\normalsize FBCSP} & 
\parbox{0.12\linewidth}{\textbf{\normalsize Deep\\ConvNet}} &
\parbox{0.12\linewidth}{\textbf{\normalsize Shallow\\ConvNet}} &
\parbox{0.12\linewidth}{\textbf{\normalsize Hybrid\\ConvNet}} &
\parbox{0.12\linewidth}{\textbf{\normalsize Residual\\ConvNet}} \\\midrule
 BCIC & 0--38 & 68.0 & +2.9 & +5.7* &  +3.6 & -0.3 \\ 
 BCIC & 4--38 & 67.8 & +2.3 & +4.1 & -1.6 & -7.0* \\ 
 HGD & 0--125 & 91.2 & +1.3 & -1.9 & +0.6 &   -2.3* \\ 
 HGD & 4--125 & 90.9 & +0.5 & +3.0* &  +1.5 &  -1.1\\ 
 Combined & 0--$f_{end}$ & 82.1 & +1.9* & +1.1 & +1.8 & -1.1 \\ 
 Combined & 4--$f_{end}$ & 81.9 & +1.2 & +3.4** & +0.3 & -3.5* \\ 
  \bottomrule
\hline
\end{tabular}
\caption{\textbf{Decoding accuracy of FBCSP baseline as well as of the deep and
shallow ConvNets}. FBCSP decoding accuracies and difference of deep and
shallow ConvNet accuracies to FBCSP results are given in percent. BCIC: BCI Competition Dataset. HGD: 
High-Gamma Dataset. Frequency range is in Hz.
Stars indicate statistically significant differences (p-values from Wilcoxon
signed-rank test, *: p \textless{} 0.05, **: p \textless{} 0.01, no
p-values were below 0.001).}
\label{table:results-fbcsp-convnets}
\end{table}

Both the deep the shallow ConvNets, with appropriate design choices (see Result \ref{subsec:results-design-choices}),
reached similar accuracies as
FBCSP-based decoding, with 
small but statistically significant
advantages for the ConvNets in some settings.
For the mean of all subjects of both datasets, accuracies of the shallow ConvNet on
0--$f_{end}$ Hz and for the deep ConvNet on 4--$f_{end}$ Hz were not statistically significantly different from 
FBCSP  (see 
Figure \ref{fig:results-fbcsp-convnets} and Table \ref{table:results-fbcsp-convnets}).
The deep ConvNet on 0--$f_{end}$ Hz and the shallow ConvNet on 4---$f_{end}$ Hz reached slightly higher 
(1.9\% 
and 3.3\% higher respectively) accuracies that were also statistically significantly different (p<0.05, Wilcoxon 
signed-rank test).
Note that all results in this section were obtained with cropped training, for a 
comparison of cropped and trial-wise training, see Section \ref{subsec:results-training-strategy}.

\begin{result}
Confusion matrices for all decoding approaches were similar \label{subsec:results-confmats}
\end{result}

\begin{figure}
\begin{subfigure}[t]{0.4\linewidth}
\centering
\includegraphics[max size={\linewidth}{0.4\paperheight}]{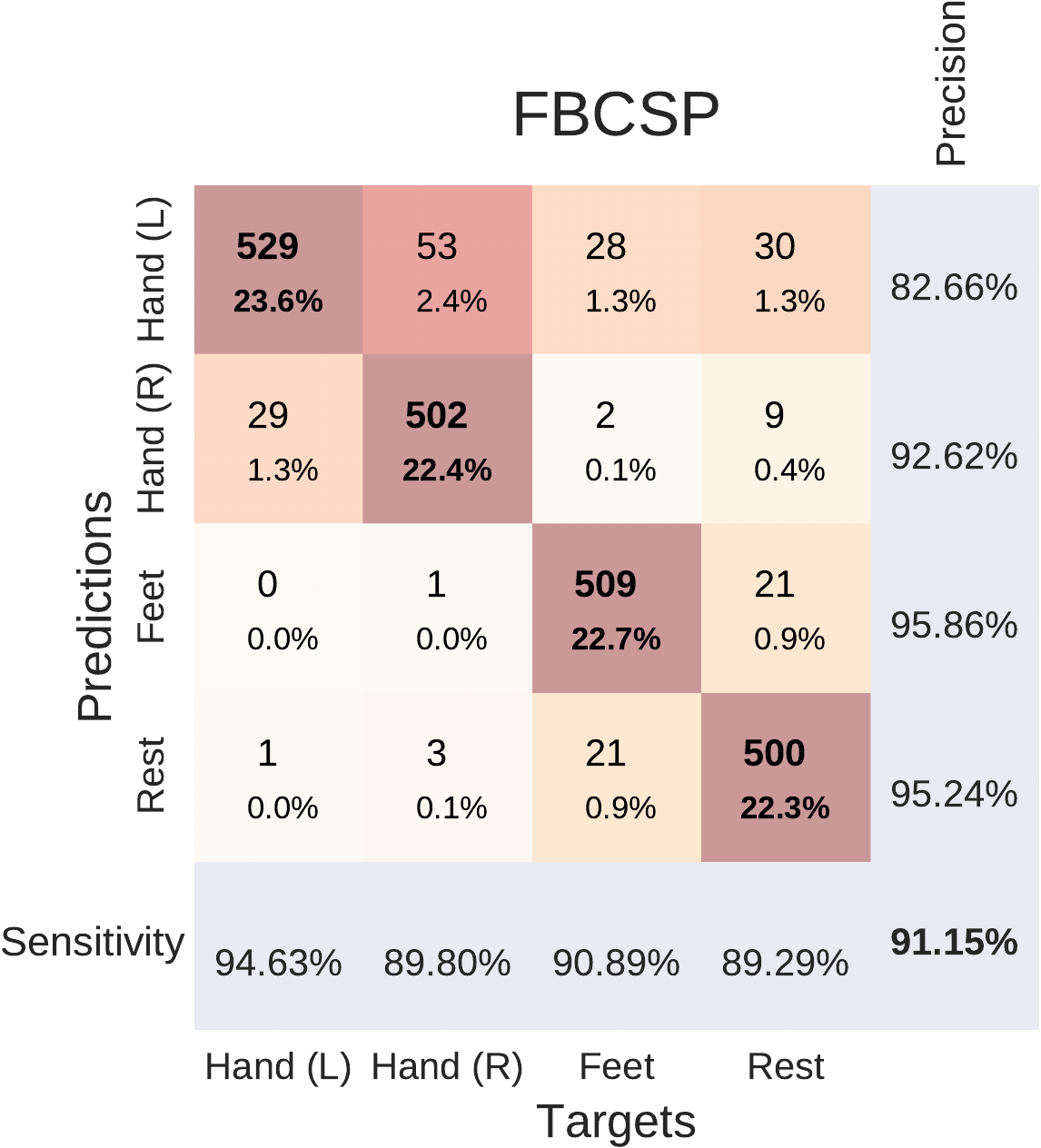}
\end{subfigure}\hspace{1cm}%
\begin{subfigure}[t]{0.4\linewidth}
\centering
\includegraphics[max size={\linewidth}{0.4\paperheight}]{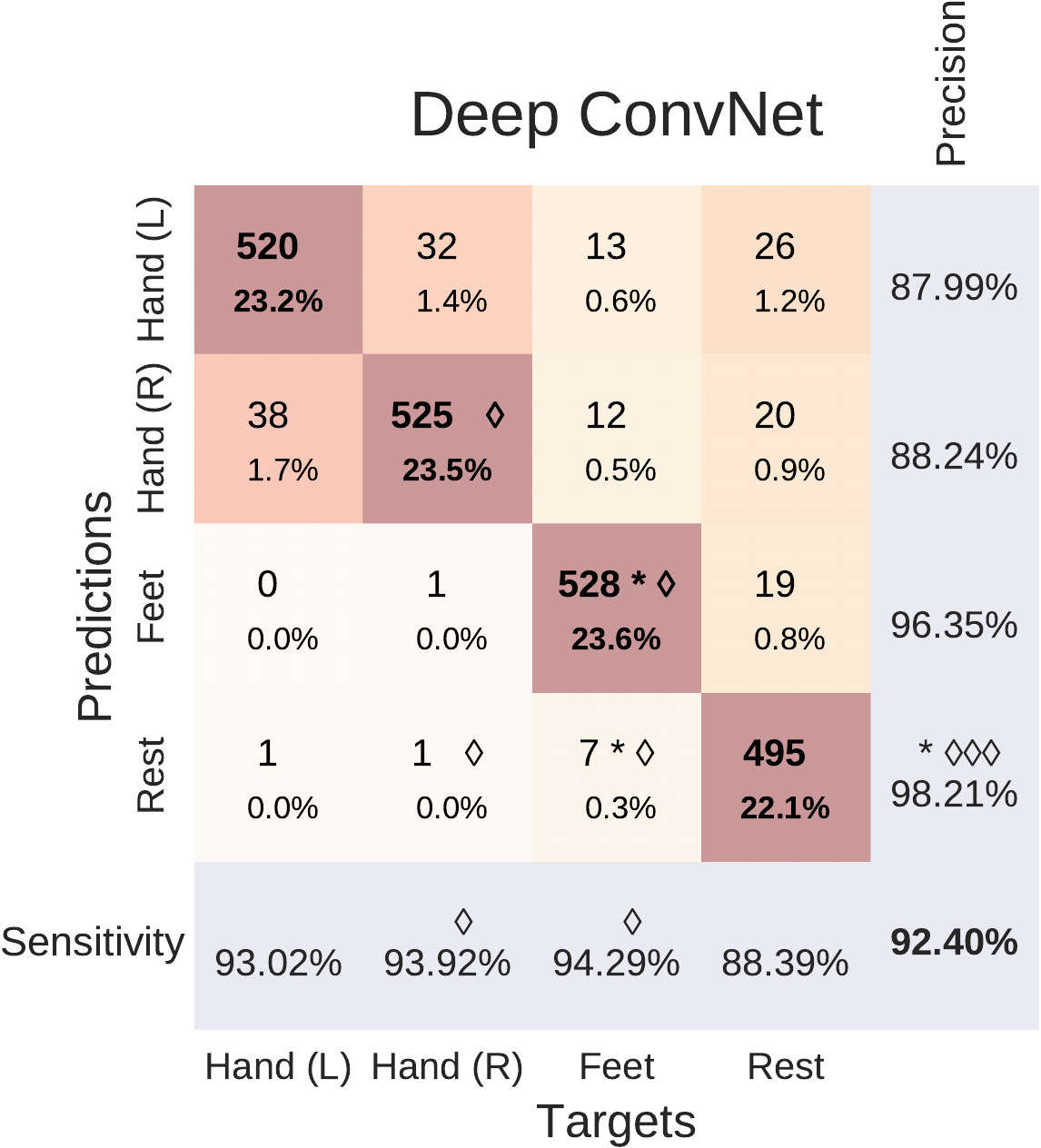}
\end{subfigure}
\begin{subfigure}[t]{0.5025\linewidth}
\centering
\includegraphics[max size={\linewidth}{0.4\paperheight}]{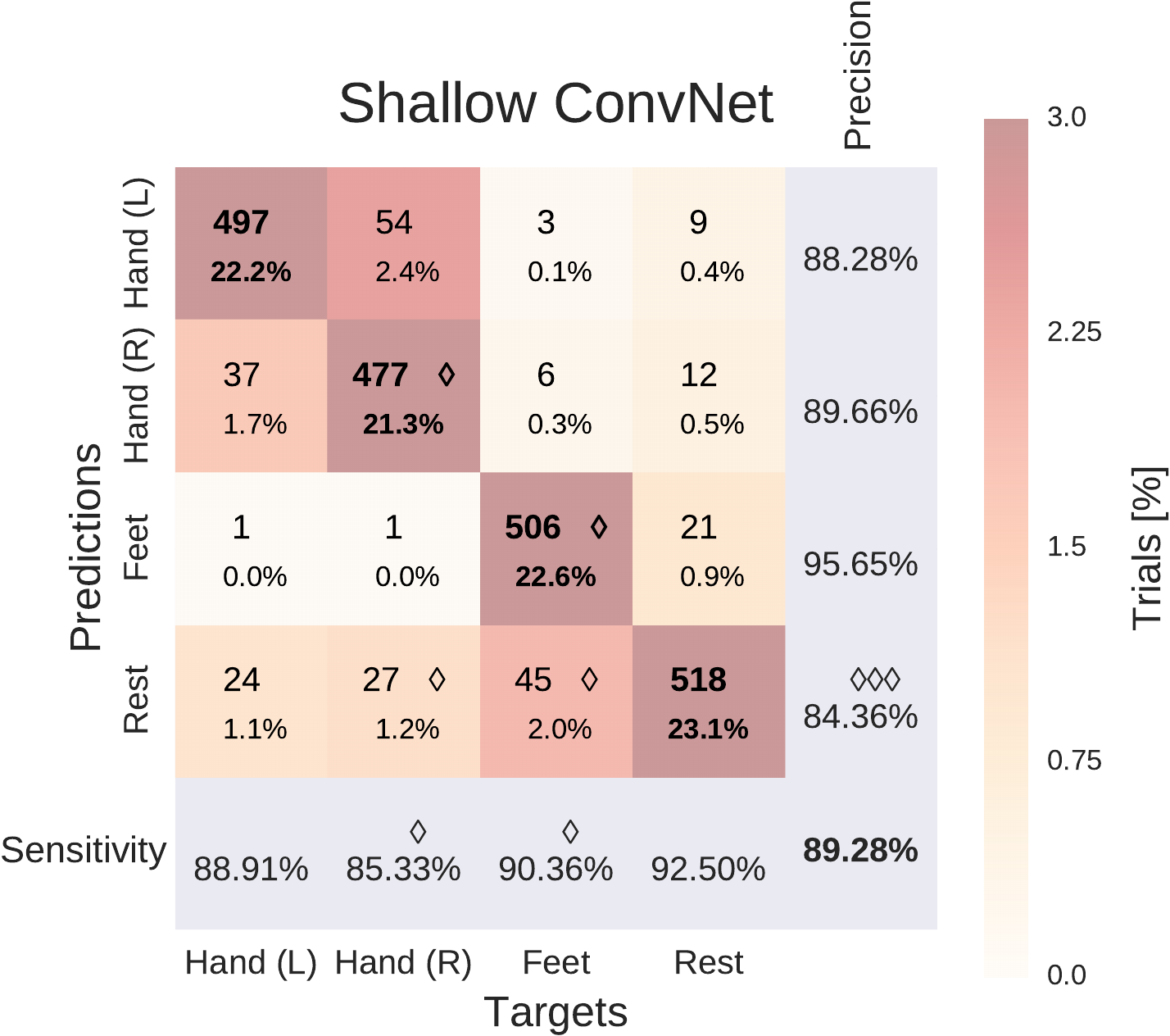}
\end{subfigure}
\caption{\textbf{Confusion matrices for FBCSP- and ConvNet-based decoding.} Results are
shown for the High-Gamma Dataset, on 0--$f_{end}$ Hz.
Each entry of row r and column c for
upper-left 4x4-square: Number of trials of target r predicted as class c
(also written in percent of all trials). Bold diagonal corresponds to
correctly predicted trials of the different classes. The lower-right
value corresponds to overall accuracy. Bottom row corresponds to
sensitivity defined as the number of trials correctly predicted for
class c / number of trials for class c. Rightmost column corresponds to
precision defined as the number of trials correctly predicted for class
r / number of trials predicted as class r. Stars indicate statistically
significantly different values of ConvNet decoding from FBCSP, diamonds
indicate statistically significantly different values between the
shallow and deep ConvNets . p<0.05: $\diamond$/*, p<0.01:
$\diamond\diamond$/**, p<0.001: $\diamond\diamond\diamond$/***, Wilcoxon signed-rank test.}
\label{fig:confmats}
\end{figure}

Confusion matrices for the High-Gamma Dataset on
0--$f_{end}$ Hz were very similar for FBCSP
and both ConvNets (see Figure \ref{fig:confmats}).
The majority of all mistakes were due to discriminating
between Hand (L) / Hand (R) and Feet / Rest, see Table \ref{table:decoding-mistakes}.
Seven entries of the confusion matrix had a statistically significant difference (p<0.05, Wilcoxon signed-rank test)
between the deep and the shallow ConvNet, in all of them the deep ConvNet performed better.
Only two differences between the deep ConvNet and FBCSP were statistically significant (p<0.05), none for the shallow 
ConvNet
and FBCSP. Confusion matrices for the BCI Competition Dataset showed a larger variability and hence a less consistent 
pattern, possibly because of the much smaller number of trials.

\begin{table}
\centering \footnotesize
\begin{tabular}{rllllll}\toprule
&
\parbox{0.12\linewidth}{\textbf{Hand (L)\\Hand (R)}} & 
\parbox{0.12\linewidth}{\textbf{Hand (L)\\Feet}} & 
\parbox{0.12\linewidth}{\textbf{Hand (L)\\Rest}} & 
\parbox{0.12\linewidth}{\textbf{Hand (R)\\Feet}} & 
\parbox{0.12\linewidth}{\textbf{Hand (R)\\Rest}} & 
\parbox{0.12\linewidth}{\textbf{Feet\\Rest}} \\\midrule
 FBCSP & 82 & 28 & 31 & 3 & 12 & 42 \\ 
 Deep & 70 & 13 & 27 & 13 & 21 & 26 \\ 
 Shallow & 99 & 3 & 34 & 5 & 37 & 73 \\ 
  \bottomrule
\hline
\end{tabular}
\caption{Decoding mistakes between class pairs. Results for the High-Gamma Dataset.
Number of trials where one class  was mistaken for the
other for each decoding method, summed per class pair. The largest
number of mistakes was between Hand(L) and Hand (R) for all three
decoding methods, the second largest between Feet and Rest (on average
across the three decoding methods). Together, these two class pairs
accounted for more than 50\% of all mistakes for all three decoding
methods. In contrast, Hand (L and R) and Feet had a small number of
mistakes irrespective of the decoding method used.}
\label{table:decoding-mistakes} 
\end{table}

\begin{result}
Hybrid ConvNets performed slightly, but statistically insignificantly, worse than deep ConvNets
\label{subsec:results-hybrid}
\end{result}

The hybrid ConvNet performed similar, but slightly worse than the deep
ConvNet, i.e., 83.8\% vs 84.0\% (p>0.5, Wilcoxon signed-rank test) on the 0--$f_{end}$-Hz dataset, 
82.1\% vs
83.1\% (p>0.9) on the
4--$f_{end}$-Hz dataset.
In both cases, the
hybrid ConvNet's accuracy was also not statistically significantly
different from FBCSP (83.8\% vs 82.1\%, p>0.4 on
0--$f_{end}$ Hz , 82.1\% vs 81.9\%, p>0.7 on 4--$f_{end}$ Hz).

\begin{result}
ConvNet design choices substantially affected decoding accuracies
\label{subsec:results-design-choices}
\end{result}

In the following, results for all design choices are reported for all
subjects from both datasets. For an overview of the different design
choices investigated, and the motivation behind these choices, we refer to
Section \ref{subsec:convnet-design-choices}.

\begin{figure}
\begin{subfigure}[t]{\linewidth}
\centering
\includegraphics[max size={\linewidth}{0.25\paperheight}]{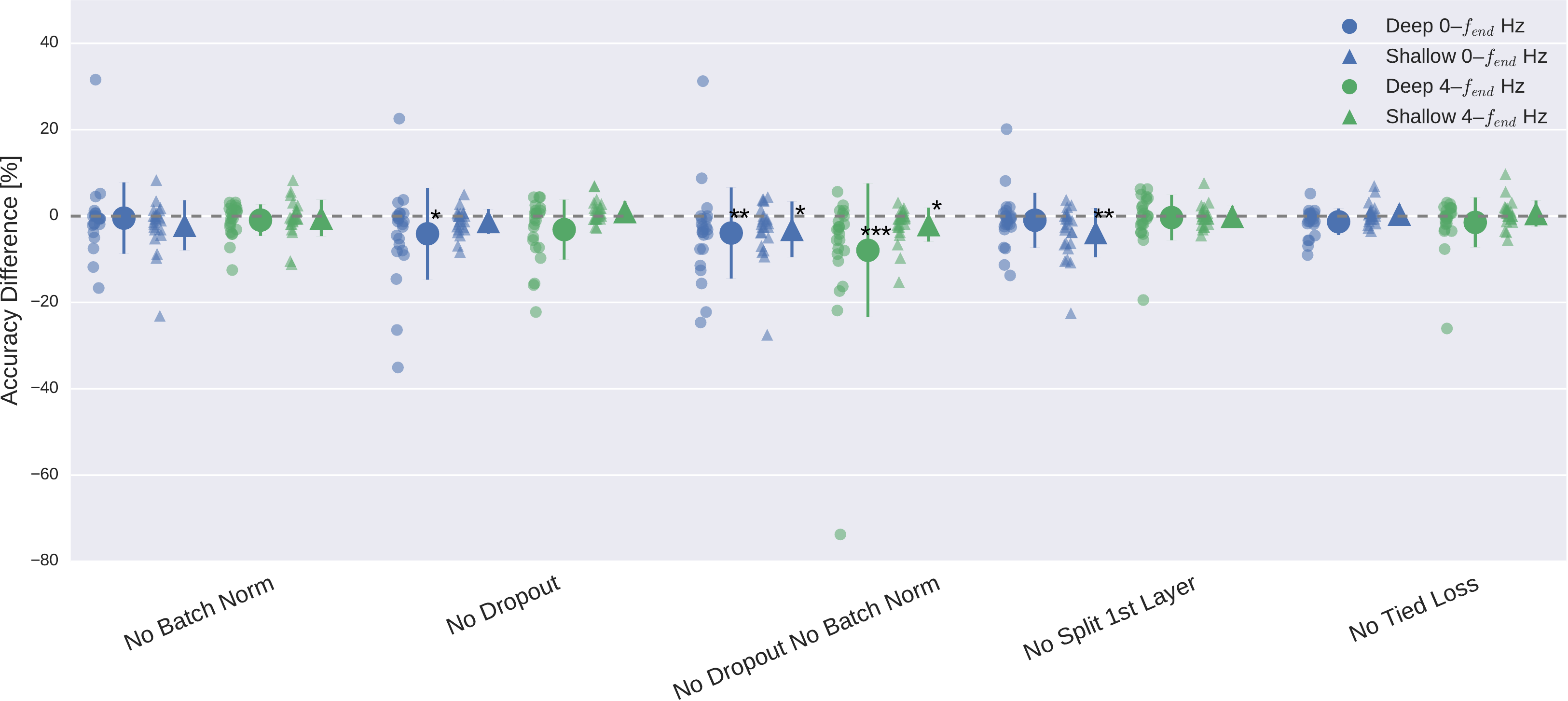}
\caption{\footnotesize{Impact of design choices applicable to both ConvNets. Shown are the
effects from the removal of one aspect from the architecture on decoding
accuracies. All statistically significant differences were accuracy
decreases. Notably, there was a clear negative effect of removing both
dropout and batch normalization, seen in both ConvNets' accuracies
and for both frequency ranges.}}
\label{fig:results-design-choices-a}
\end{subfigure}
\begin{subfigure}[t]{\linewidth}
\centering
\includegraphics[max size={\linewidth}{0.25\paperheight}]{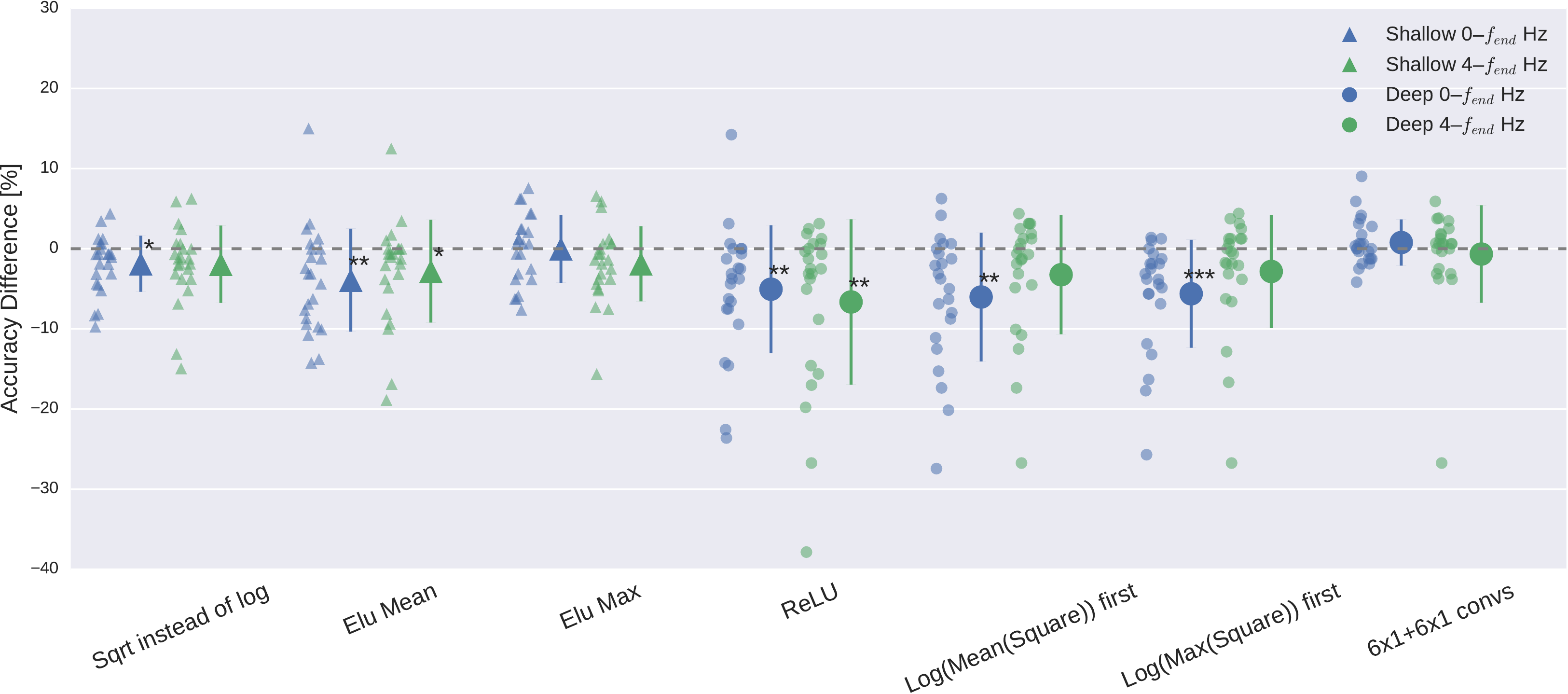}
\caption{\footnotesize{Impact of different types of nonlinearities, pooling modes and
filter sizes. Results are given independently for the deep ConvNet and
the shallow ConvNet. As before, all statistically significant
differences were from accuracy decreases. Notably, replacing ELU by ReLU
as nonlinearity led to decreases on both frequency ranges, which were
both statistically significant.}}
\label{fig:results-design-choices-b}
\end{subfigure}
\caption{\textbf{Impact of ConvNet design choices on decoding accuracy.}
Accuracy differences of baseline and design choices on x-axis for the 0--$f_{end}$-Hz and 4--$f_{end}$-Hz datasets.
Each small marker represents accuracy difference for one subject, each larger marker represents mean accuracy 
difference across all subjects of both datasets.
Bars: standard error of the differences across subjects. Stars indicate
statistically significant differences to baseline (Wilcoxon signed-rank test, p\textless{}0.05: *, p\textless{}0.01: **,
p\textless{}0.001=***)}
\label{fig:results-design-choices}
\end{figure}

Batch normalization and dropout significantly increased accuracies.This
became especially clear when omitting both simultaneously (see Figure \ref{fig:results-design-choices-a}).
Batch normalization provided a larger accuracy increase for the shallow ConvNet, whereas dropout provided a larger 
increase for the deep ConvNet.
For both networks and for both frequency bands, the only statistically significant accuracy differences were accuracy 
decreases after removing dropout for the deep ConvNet on 0--$f_{end}$-Hz data or removing batch normalization and 
dropout 
for both networks and frequency ranges (p<0.05, Wilcoxon signed-rank test).
Usage of tied loss did not affect the accuracies very much, never yielding statistically significant differences
(p>0.05).
Splitting the first layer into two convolutions had the strongest accuracy increase on the 0--$f_{end}$-Hz data for the 
shallow ConvNet, where it is also the only statistically significant difference (p<0.01).

For the deep ConvNet, using ReLU instead of ELU as nonlinearity in all
layers worsened performance (p<0.01, see Figure \ref{fig:results-design-choices-b} on the right side).
Replacing the 10x1 convolutions by 6x1+6x1 convolutions did not statistically significantly affect the performance 
(p>0.4).

\begin{result}
Recent deep learning advances substantially increased accuracies
\label{subsec:results-recent}
\end{result}

\begin{figure}
\centering
\includegraphics[max size={1\linewidth}{0.56\paperheight}]{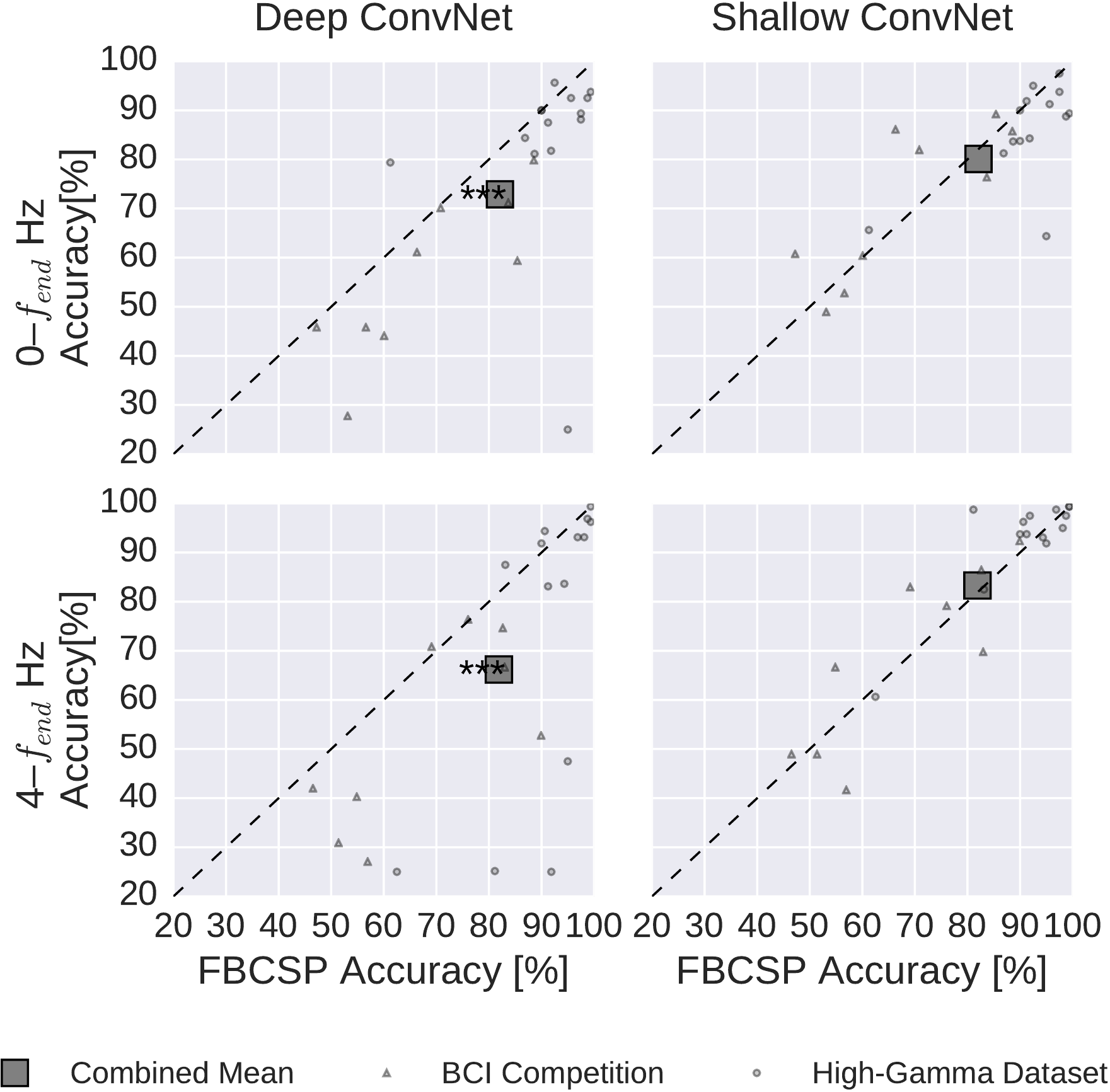}
\caption{\textbf{Impact of recent advances on overall decoding accuracies.}
Accuracies without batch normalization, dropout and ELUs.
All conventions as in Figure \ref{fig:results-fbcsp-convnets}.
In contrast to the results on Figure \ref{fig:results-fbcsp-convnets}, the deep ConvNet without implementation of these 
recent methodological advances performed worse than FBCSP; the difference was statistically significant for both 
frequency ranges.}
\label{fig:results-recent}
\end{figure}

Figure \ref{fig:results-recent} clearly shows that only recent advances in deep learning methods together (by which we
mean the combination of batch normalization, dropout and ELUs) allowed our deep ConvNet to be competitive with FBCSP.
Without these recent advances, the deep ConvNet had statistically significantly worse accuracies than FBCSP for both 
0--$f_{end}$-Hz and 4--$f_{end}$-Hz data (p<0.001, Wilcoxon signed-rank test).
The shallow ConvNet was less strongly affected, with no statistically significant accuracy difference to FBCSP (p>0.2).

\begin{result}
Residual network performed worse than deep ConvNet
\label{subsec:results-resnet}
\end{result}

\begin{table}
\centering 
\begin{tabular}{lllll}\toprule
\textbf{Dataset} & 
\parbox{0.15\linewidth}{\textbf{\normalsize Frequency\\range [Hz]}} & 
{\textbf{Accuracy}} & 
\parbox{0.15\linewidth}{\textbf{Difference\\to deep}} &
{\textbf{p-value}} \\\midrule
 BCIC & 0--38 & 67.7 & -3.2 & 0.13 \\ 
 BCIC & 4--38 &  60.8 & -9.3 & 0.004** \\
HGD & 0--125 & 88.9 & -3.5 & 0.020* \\
HGD & 4--125 & 89.8 & -1.6 & 0.54 \\
Combined & 0--$f_{end}$ & 80.6 & -3.4 & 0.004** \\
Combined & 4--$f_{end}$ & 78.5 & -4.9 & 0.01* \\
  \bottomrule
\hline
\end{tabular}
\caption{\textbf{Decoding accuracies residual networks and difference to deep ConvNets.}
BCIC: BCI Competition Dataset. HGD: High-Gamma Dataset.
Accuracy is mean accuracy in percent.
P-value from Wilcoxon signed-rank test for the statistical significance of the differences to
the deep ConvNet (cropped training).
Accuracies were always slightly worse than deep ConvNet, statistically significantly different for both frequency 
ranges on the combined dataset.}
\label{table:results-resnet}
\end{table}

Residual networks had consistently worse accuracies than the deep
ConvNet as seen in Table \ref{table:results-resnet}.
All accuracies were lower and the difference was statistically significant for both frequency ranges on the combined 
dataset.

\subsection{Training Strategy}\label{subsec:results-training-strategy}
\begin{result}
Cropped training strategy improved deep ConvNet on higher frequencies
\label{subsec:results-cropped}
\end{result}

\begin{figure}
\centering
\includegraphics[max size={1\linewidth}{0.4\paperheight}]{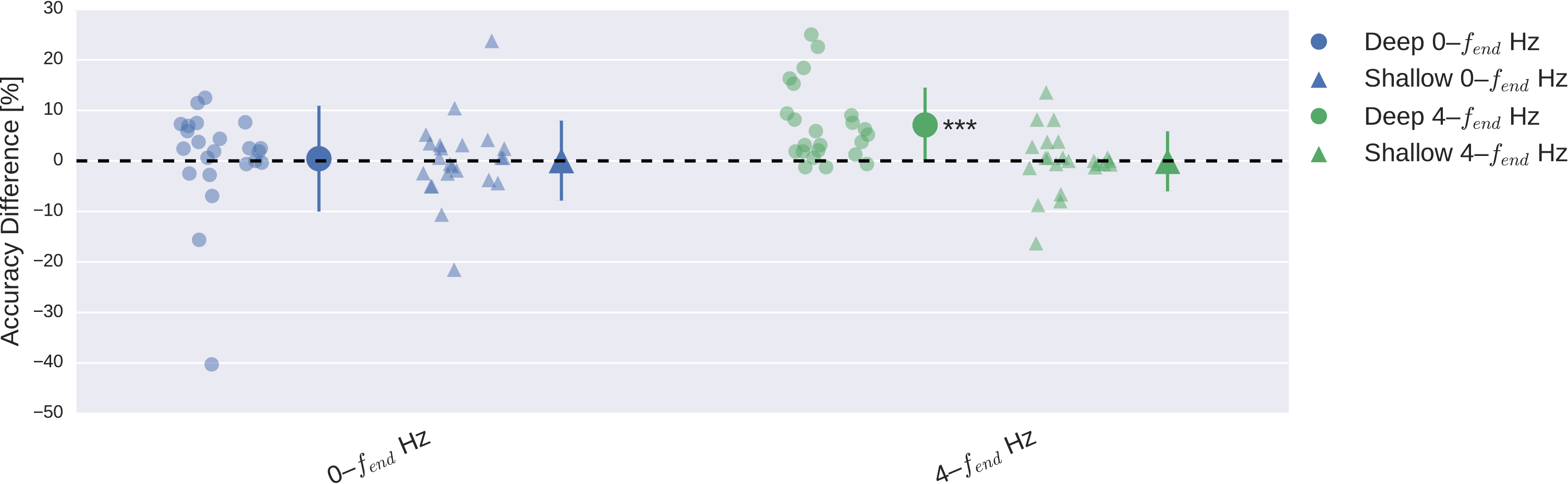}
\caption{\textbf{Impact of training strategy (cropped vs trial-wise training) on
accuracy.} Accuracy difference for both frequency ranges and both
ConvNets when using cropped training instead of trial-wise training.
Other conventions as in Figure \ref{fig:results-design-choices}.
Cropped training led to better accuracies for almost 
all
subjects for the deep ConvNet on the
4--$f_{end}$-Hz frequency range.}
\label{fig:results-cropped}
\end{figure}

Cropped training increased accuracies statistically significantly for
the deep ConvNet on the 4--$f_{end}$-Hz data (p<1e-5, Wilcoxon signed-rank test).
In all other settings (0--$f_{end}$-Hz data,
shallow ConvNet), the accuracy differences were not statistically
significant (p>0.1) and showed a lot of variation between subjects.

\begin{result}
Training ConvNets took substantially longer than FBCSP
\label{subsec:results-times}
\end{result}

FBCSP was substantially faster to train than the ConvNets with cropped training, by a factor
of 27--45 on the BCI Competition Dataset and a factor of 5--9 on the High-Gamma Dataset.
Training times are end-to-end, i.e., include the loading and preprocessing of the data.
These times are only meant to give a rough estimate of the training times as there were differences in the computing 
environment between ConvNets training and FBCSP training.
Most importantly, FBCSP was trained on CPU, while the networks were trained on GPUs (see Section 
\ref{subsec:software-hardware}).
Longer relative training times for FBCSP on the High-Gamma Dataset can be explained by the larger number of frequency 
bands we use on the High-Gamma Dataset.
Online application of the trained ConvNets does not suffer from the same speed disadvantage compared to FBCSP; the fast
prediction speed of trained ConvNets make them well suited for decoding in real-time BCI applications.

\begin{table}
\centering 
\begin{tabular}{lllllll}\toprule
\textbf{Dataset} & 
\parbox{0.1\linewidth}{\textbf{FBCSP}} &
\textbf{std} &
\parbox{0.1\linewidth}{\textbf{Deep\\ConvNet}} &
\textbf{std} &
\parbox{0.1\linewidth}{\textbf{Shallow\\ConvNet}} &
\textbf{std} \\ \midrule
BCIC & 00:00:33 & <00:00:01 & 00:24:46 & 00:06:01 &  00:15:07 & 00:02:54 \\ 
HGD & 00:06:40 & 00:00:54 & 1:00:40 & 00:27:43 & 00:34:25 & 00:16:40 \\
  \bottomrule
\hline
\end{tabular}
\caption{
\textbf{Training times.}
Mean times across subjects given in Hours:Minutes:Seconds.
BCIC: BCI Competition Dataset. HGD: High-Gamma Dataset.
Std is standard deviation across subjects.
ConvNets take substantially longer to train than FBCSP, especially the deep ConvNet.}
\label{table:results-times}
\end{table}

\subsection{Visualization}\label{subsec:results-visualization}

\begin{result}
Band power topographies show event-related ``desynchronization/synchronization'' typical for motor tasks
\label{subsec:results-spectral-topo}
\end{result}

\begin{figure}
\centering
\includegraphics[max size={1\linewidth}{0.4\paperheight}]{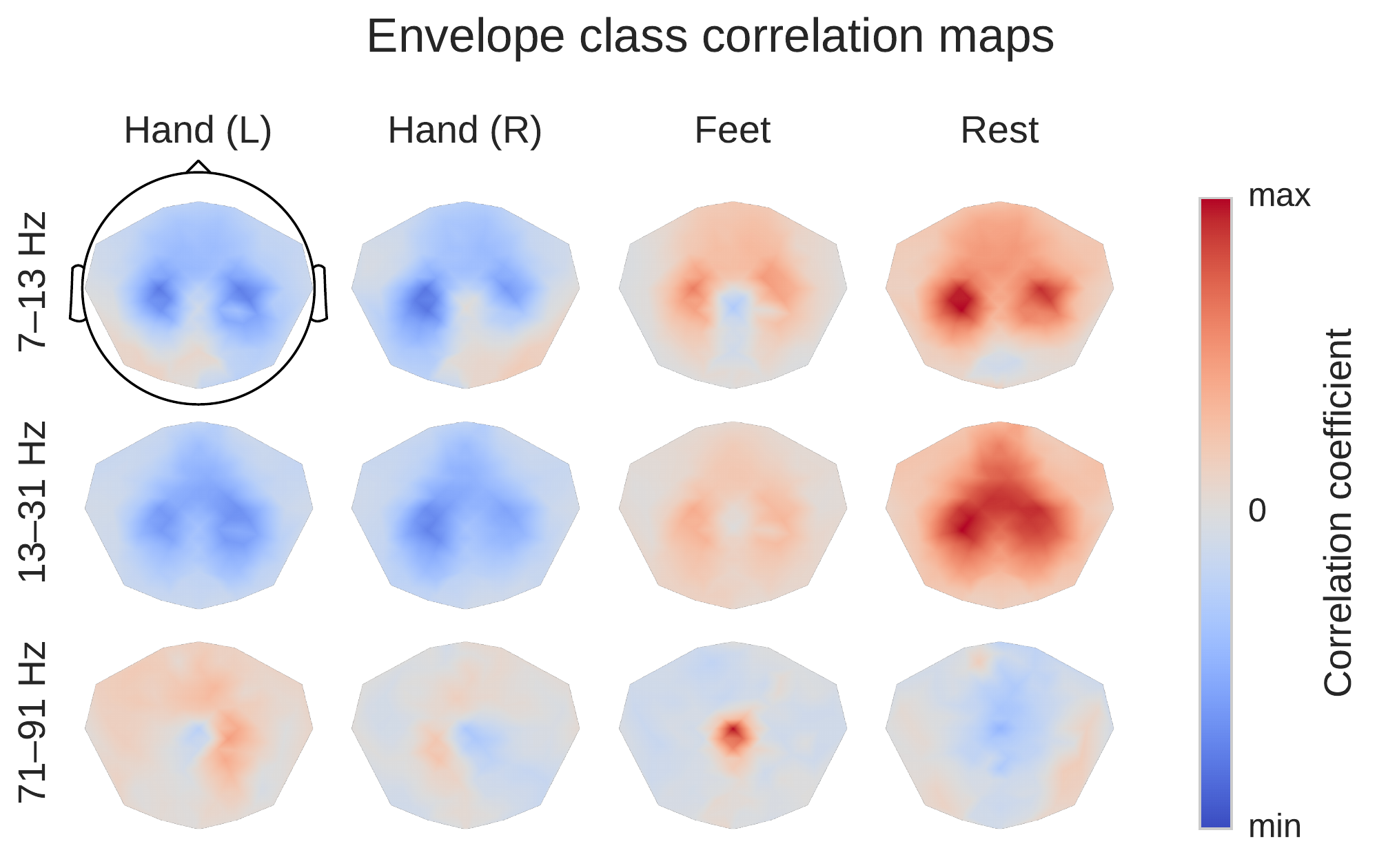}
\caption{\textbf{Envelope-class correlations for alpha, beta and gamma bands for all
classes.}
Average over subjects from the High-Gamma Dataset. Colormaps
are scaled per frequency band/column. This is a ConvNet-independent visualization, for an explanation of the 
computation see Section \ref{subsec:correlation-visualization}. Scalp plots show spatial distributions of 
class-related spectral amplitude changes well in line with the literature.}
\label{fig:results-spectral-topo}
\end{figure}

Before moving to ConvNet visualization, we examined the spectral
amplitude changes associated with the different movement classes in the
alpha, beta and gamma frequency bands, finding the expected overall
scalp topographies (see Figure \ref{fig:results-spectral-topo}).
For example, for the alpha (7--13 Hz) frequency band, there was a
class-related power decrease (anti-correlation in the class-envelope
correlations) in the left and right pericentral regions with respect to
the hand classes, stronger contralaterally to the side of the hand
movement , i.e., the regions with pronounced power decreases lie around
the primary sensorimotor hand representation areas. For the feet class,
there was a power decrease located around the vertex, i.e., 
approx. above the primary motor foot area. As expected, opposite changes (power
increases) with a similar topography were visible for the gamma band
(71--91 Hz).

\begin{result}
Input-feature unit-output correlation maps show learning progression through the ConvNet layers
\label{results-correlation}
\end{result}

\begin{figure}
\centering
\includegraphics[max size={1\linewidth}{0.4\paperheight}]{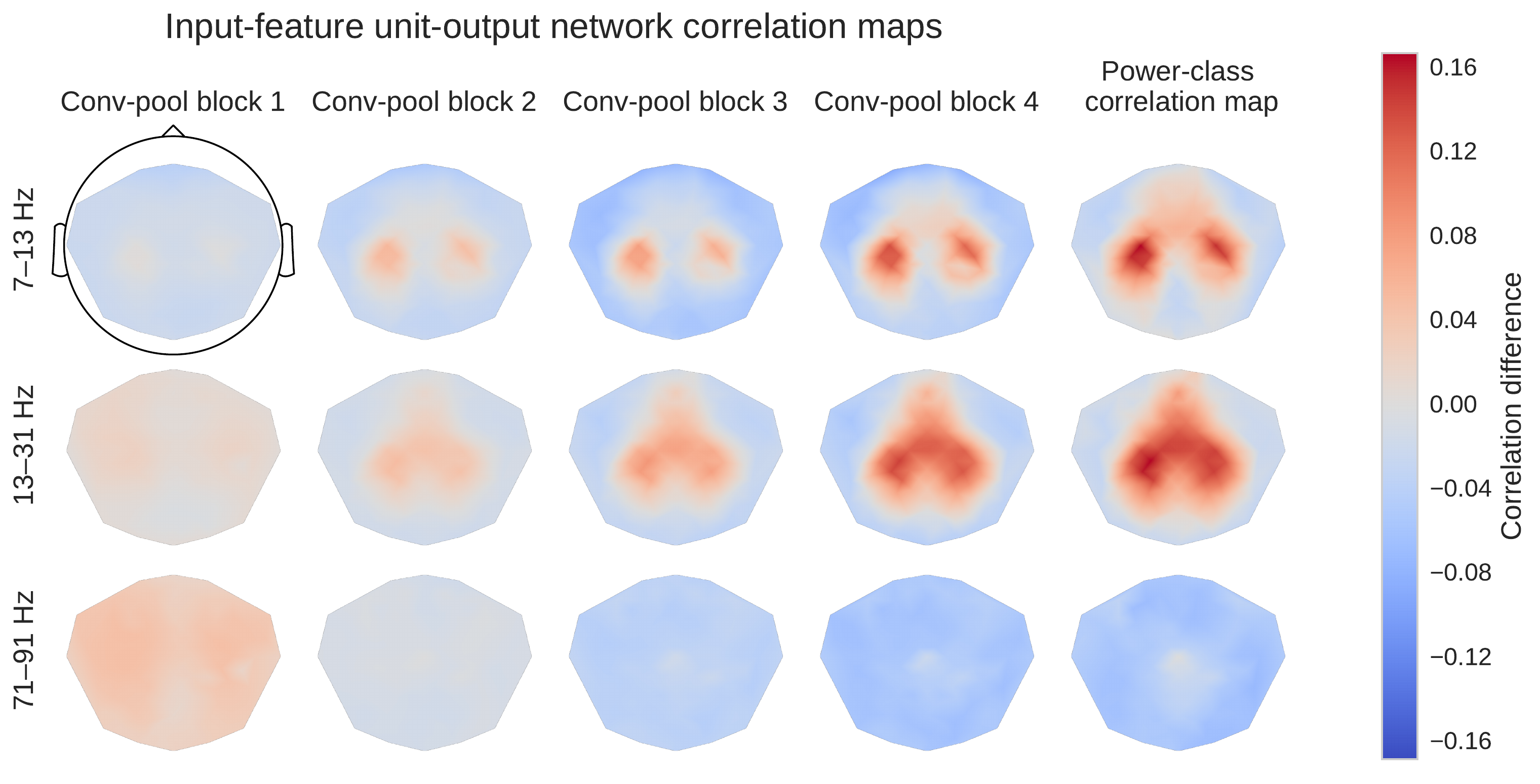}
\caption{\textbf{Power input-feature unit-output network correlation maps for all conv-pool blocks of the deep ConvNet.}
Correlation difference indicates the difference of correlation coefficients obtained with the trained and untrained 
model for each electrode respectively and is visualized as a topographic scalp plot.
Details see Section \ref{subsec:correlation-visualization}.
Rightmost column shows the correlation between the envelope of the EEG signals in each of the three analyzed frequency 
bands and the four classes.
Notably, the absolute values of the correlation differences became
larger in the deeper layers and converged to patterns that were very
similar to those obtained from the power-class correlations.}
\label{fig:results-correlation}
\end{figure}

\begin{figure}
\centering
\includegraphics[max size={1\linewidth}{0.2\paperheight}]{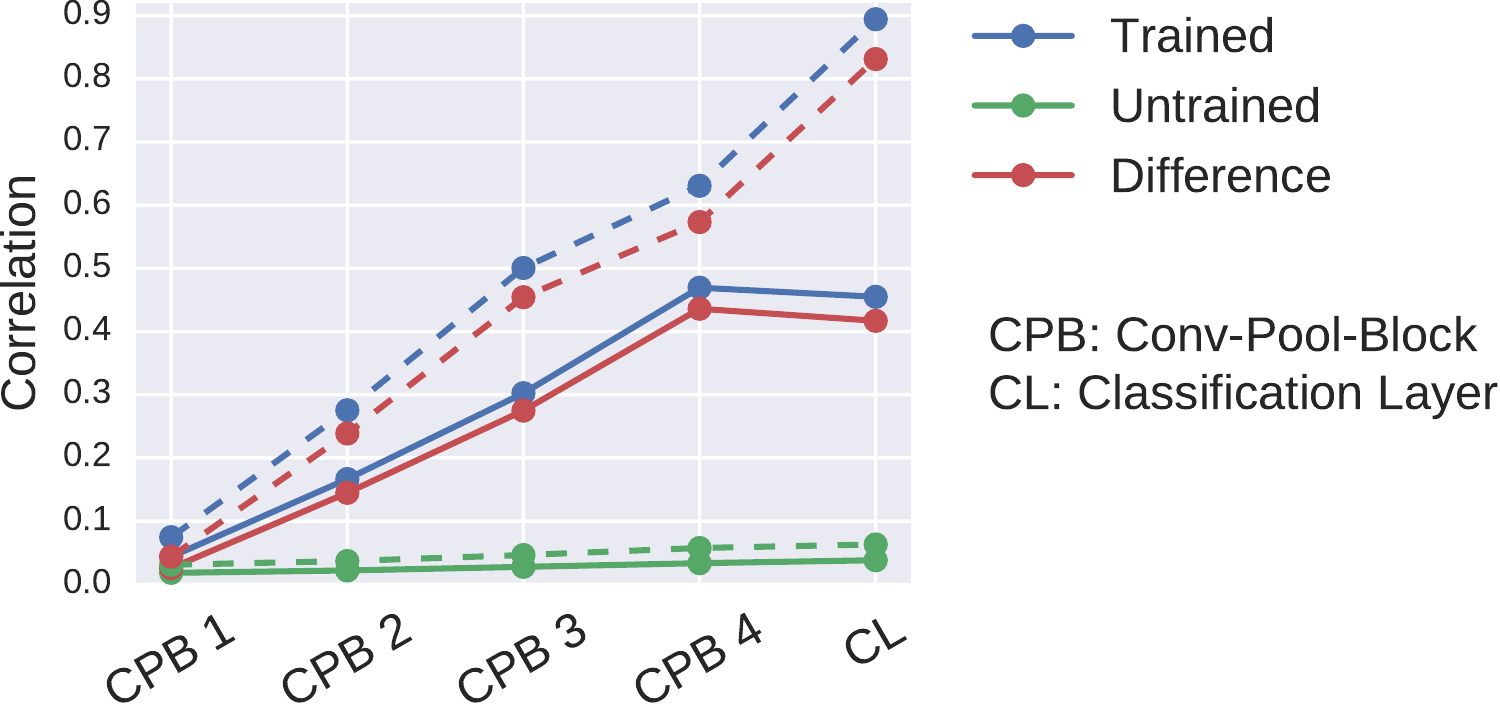}
\caption{\textbf{Absolute correlations between unit outputs and class labels.}
Each dot
represents absolute correlation coefficients for one layer of the deep ConvNet.
Solid lines
indicate result of taking mean over absolute correlation coefficients
between classes and filters. Dashed lines indicate result of first
taking the maximum absolute correlation coefficient per class (maximum
over filters) and then the mean over classes. Absolute correlations
increased almost linearly with increasing depth of the layer.}
\label{fig:results-correlation-unit-outs-labels}
\end{figure}

We used our input-feature unit-output correlation mapping technique to
examine the question how correlations between EEG power and the
behavioral classes are learnt by the network. Figure
\ref{fig:results-correlation} shows the input-feature
unit-output correlation maps for all four conv-pooling-blocks of the
deep ConvNet, for the group of subjects of the High-Gamma Dataset. As a
comparison, the Figure also contains the correlation between the power
and the classes themselves as described in Section \ref{subsec:correlation-visualization}.
The differences of the absolute correlations show which regions were more correlated with
the unit outputs of the trained ConvNet than with the unit outputs of
the untrained ConvNet; these correlations are naturally undirected.
Overall, the input-feature unit-output correlation maps became more
similar to the power-class correlation maps with increasing layer depth.
This gradual progression was also reflected in an increasing correlation
of the unit outputs with the class labels with increasing depth of the
layer (see Figure \ref{fig:results-correlation-unit-outs-labels}).

\begin{result}
Input-perturbation network-prediction correlation maps show causal effect of spatially localized band power features on 
ConvNet predictions
\label{results-perturbation}
\end{result}

We show three visualizations extracted from input-perturbation
network-prediction correlations, the first two to show the frequency
profile of the causal effects, the third to show their topography.

\begin{figure}
\centering
\includegraphics[max size={1\linewidth}{0.2\paperheight}]{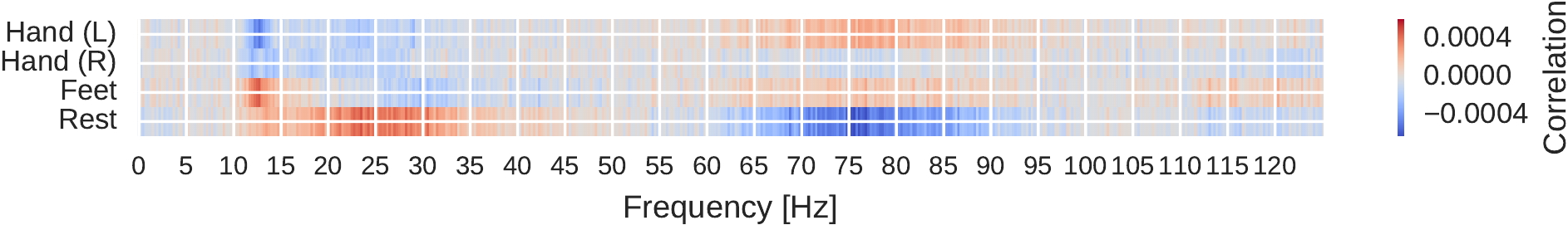}
\caption{\textbf{Input-perturbation network-prediction correlations for all frequencies
for the deep ConvNet, per class.}
Plausible correlations, e.g., rest
positively, other classes negatively correlated with the amplitude changes in frequency range from 20 Hz to 30 Hz.}
\label{fig:results-perturbation-matrix}
\end{figure}

Thus, first, we computed the mean across electrodes for each class
separately to show correlations between classes and frequency bands. We
see plausible results, for example, for the rest class, positive
correlations in the alpha and beta bands and negative correlations in
the gamma band (see Figure \ref{fig:results-perturbation-matrix}).

\begin{figure}
\centering
\includegraphics[max size={1\linewidth}{0.2\paperheight}]{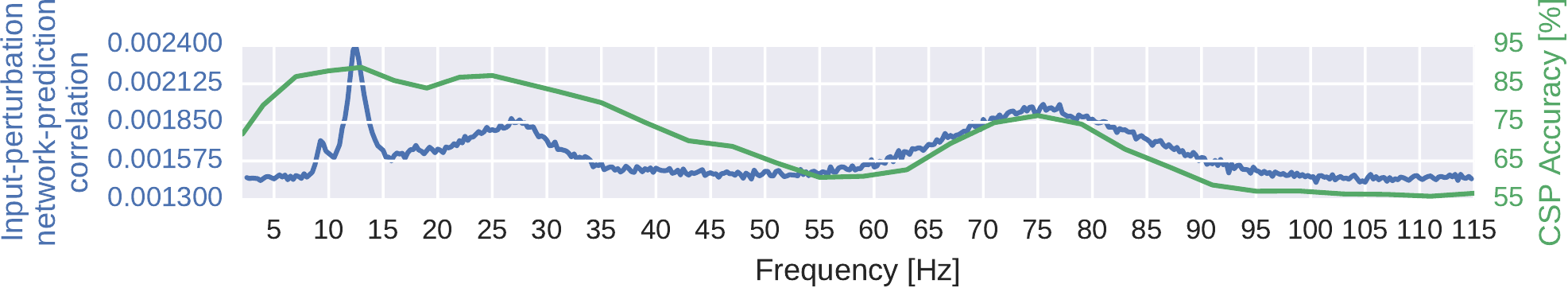}
\caption{\textbf{Absolute input-perturbation network-prediction correlation frequency
profile for the deep ConvNet.}
Mean absolute correlation value across
classes. CSP binary decoding accuracies for different frequency bands
for comparison, averaged across subjects and class pairs. Peaks in
alpha, beta and gamma band for input-perturbation network-prediction
correlations and CSP accuracies.}
\label{fig:results-perturbation-function}
\end{figure}

Then, second, by taking the mean of the absolute values both over all
classes and electrodes, we computed a general frequency profile. This
showed clear peaks in the alpha, beta and gamma bands (see Figure
\ref{fig:results-perturbation-function}).
Similar peaks were seen in the means of the CSP binary decoding accuracies for the same frequency range.

\begin{figure}
\centering
\includegraphics[max size={1\linewidth}{0.4\paperheight}]{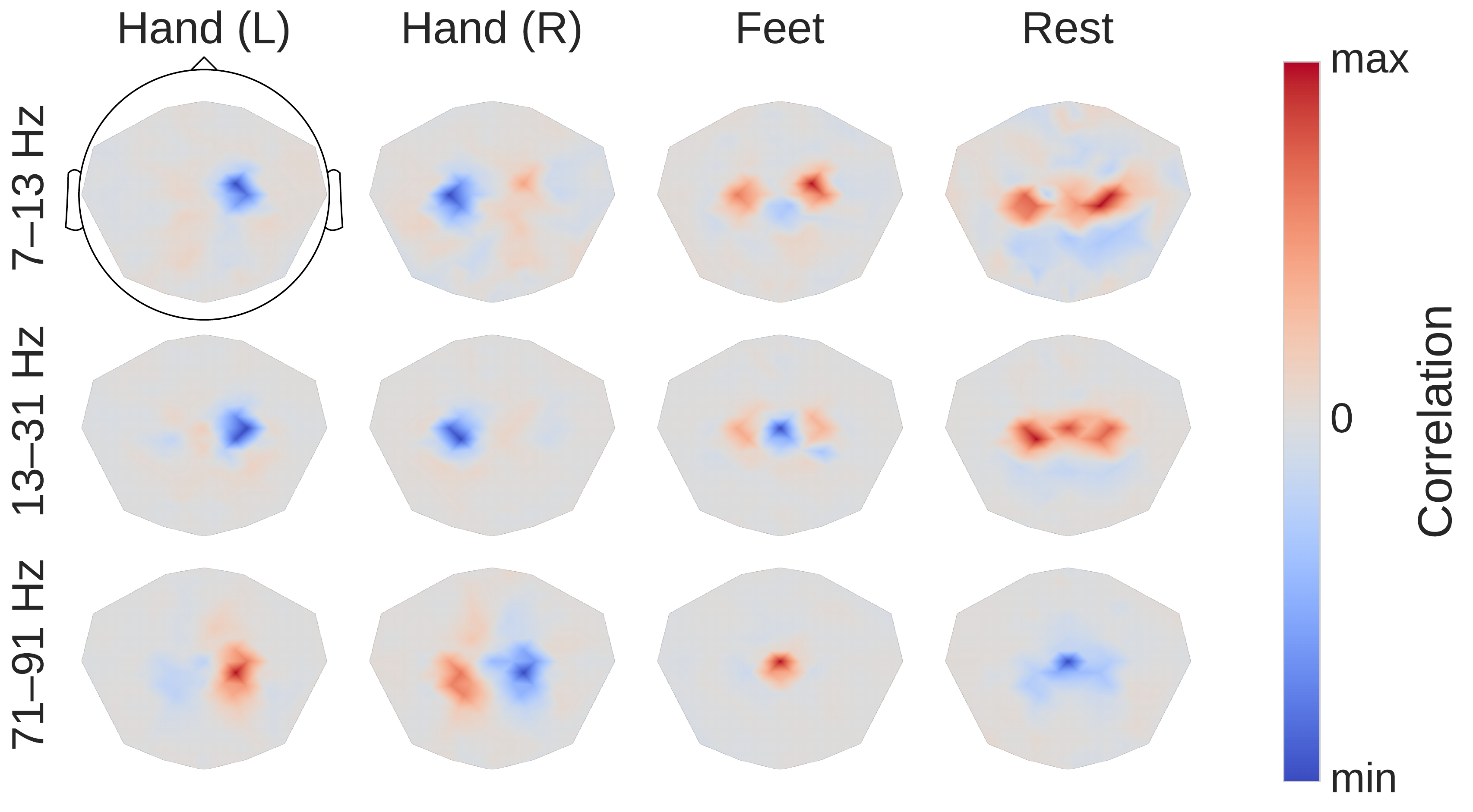}
\caption{\textbf{Input-perturbation network-prediction correlation maps for the deep
ConvNet.}
Correlation of class predictions and amplitude changes.
Averaged over all subjects of the High-Gamma Dataset.
Colormaps are scaled per scalp plot. Plausible scalp maps for all
frequency bands, e.g. contralateral positive correlations for the hand
classes in the gamma band.}
\label{fig:results-perturbation-topo}
\end{figure}

Thirdly, scalp maps of the input-perturbation effects on network
predictions for the different frequency bands, as shown in Figure \ref{fig:results-perturbation-topo}, 
show spatial distributions expected for motor tasks in the alpha, beta and --- for the first time for such a non-invasive EEG decoding visualization --- for the high gamma band.
These scalp maps directly reflect the behavior of the ConvNets and one needs to be careful when making inferences about 
the data from them.
For example, the positive correlation on the right side of the scalp for the Hand (R) class in the alpha band only 
means the ConvNet increased its prediction when the amplitude at these electrodes was increased independently of other 
frequency bands and electrodes.
It does not imply that there was an increase of amplitude for the right hand class in 
the data. Rather, this correlation could be explained by the ConvNet reducing common noise between both locations, for 
more explanations of these effects in case of linear models see \citet{haufe_interpretation_2014}.
Nevertheless, for the first time in non-invasive EEG, these maps clearly revealed the global somatotopic organization of causal contributions of motor cortical gamma band activity to decoding right and left hand as well as foot movements.

In summary, our visualization methods proved useful to map the spatial distribution of the features learned by the ConvNets to perform single-trial decoding of the different movement classes and in different physiologically important frequency bands. 
\section{Discussion}\label{sec:discussion}

This study systematically evaluated ConvNet of different architectures and with different design choices against a 
validated baseline method, i.e. FBCSP.
This study shows that ConvNets allow accurate motor decoding from EEG, that 
recent deep-learning techniques are critical to boost ConvNet performance, and that a cropped ConvNet training strategy 
can further increase decoding performance.
Thus, ConvNets can achieve successful end-to-end learning from EEG with just 
minimal preprocessing. This study also demonstrates that novel ConvNets visualization offer new possibilities in brain 
mapping of informative EEG features.

\subsection{Architectures and design choices}
\subsubsection{ConvNets vs. FBCSP}
Our results demonstrate that deep and shallow
ConvNets, with appropriate design choices, are able to --- at least --- reach the accuracies of
FBCSP for motor decoding from EEG (see Result \ref{subsec:results-fbcsp-convnets}).
In our main comparison for the combined datasets (see Table \ref{table:results-fbcsp-convnets}), the accuracies
of both deep and shallow ConvNets are very close and slightly higher than the accuracies of 
FBCSP.
As filter bank common spatial patterns is the de facto standard for motor decoding from EEG recordings, this strongly 
implies ConvNets are also a suitable method for motor decoding.
While we have shown deep ConvNets to be competitive
with standard FBCSP, a lot of variants of FBCSP exist. For example, many
regularized variants of CSP exist that can be used inside FBCSP \citep{lotte_regularizing_2011,samek_robust_2014};
a comparison to these could further show the exact tradeoff between the more generic 
ConvNets and the more domain-specific FBCSP.

\subsubsection{Role of recent deep learning advances}
Success depends on using recent developments in deep learning.
The accuracy increase that we demonstrate when using batch normalization, dropout 
and exponential linear units implies that general advances in deep learning can also improve brain-signal
decoding.
The improvement from using these techniques replicates recent findings in computer vision and other fields.
In our study, improvements were most pronounced for the deep ConvNet on 4--$f_{end}$Hz data (see Result 
\ref{subsec:results-recent}), indicating that the networks can easily overfit in this setting,
where band power features are likely dominant.
This is consistent with our observation that cropped training, which combats overfitting by increasing the 
number of training examples, also drastically increased accuracies on 4--$f_{end}$Hz data (see Result 
\ref{subsec:results-cropped}).
There seemed to be some further gains when combining both batch normalization and dropout, albeit with some 
variation across architectures and frequency bands.
This improvement was not clear from the start as batch normalization can in some cases remove the need for dropout 
\citep{ioffe_batch_2015}, however this improvement was also found in another study using ConvNets to decode EEG data 
\citep{lawhern_eegnet:_2016}.
The smaller improvement batch normalization yielded for the deep ConvNet is consistent with the claim that ELUs 
already allow fast learning \citep{clevert_fast_2016}.
However, all of these findings are limited by the fact that there can be interactions between these methods and with 
all other hyperparameters.
As of yet, we also do not have a clear explanation for the large difference in accuracies obtained with ReLUs compared 
to ELUs; a recent study on computer vision tasks did not find these differences \citep{mishkin_systematic_2016}. 
Mathematically and empirically analyzing the behavior of ELUs and ReLUs for oscillatory signals and typical EEG noise 
might shed some light on plausible reasons.

\subsubsection{ConvNet architectures and interactions with discriminative features}

Another finding of our study was that the shallow
ConvNets performed as good as the deep ConvNets, in contrast to
the hybrid and residual architectures (see Results \ref{subsec:results-fbcsp-convnets}, 
\ref{subsec:results-hybrid} and \ref{subsec:results-resnet}).
These observations could possibly be better understood by investigating more
closely what discriminative features there are in the EEG data and what
architectures can hence best use them. For example, it would be
interesting to study the effect of more layers when the networks use
mostly EEG band power features, phase-related features, or a combination
thereof (c.f. \citet{hammer_role_2013}, for the role of power and phase in motor decoding) and whether there are 
features for which a deeper hierarchical representation could be beneficial.

We observed that squaring was important for the shallow but not for the deep ConvNet (see Result  
\ref{subsec:results-design-choices}).
The worse performance of the shallow ConvNet with ELU instead of squaring may be explained as follows.
Squaring naturally allows the network to more easily extract band power features:
In combination with the approximately zero-mean input, the network would already capture the signal's variance by 
squaring.
To see this, assume that the two bandpass-filter-like and spatial-filter-like convolutional layers extract an 
oscillatory source in a specific frequency band; the squaring and mean pooling then directly computes the
variance of this source in the pool regions.
With ELUs instead of squaring, the positive parts of the oscillation would remain unchanged while the negative ones 
would be suppressed; the mean of the pool region would still be larger for larger amplitudes of the oscillation, but 
less strongly so than for the square activation.
The effects of ELU and squaring for the deep ConvNet are less straightforward to analyze, since the pooling regions 
in our deep ConvNet were much smaller than for the shallow ConvNet (3 vs 75 samples) and might 
thus not cover a large enough time span to compute a very robust and informative variance average.

\subsubsection{Possibilities for substantial decoding accuracy improvements?}
In the analyses presented here, ConvNets did not improve accuracies over FBCSP by a large margin. 
Significant improvements, if present, were never larger than 3.5 percent on the combined dataset with a lot of 
variation per subject (see Result \ref{subsec:results-fbcsp-convnets}).
However, the deep ConvNets as used here may have learned features different from FBCSP, which could explain their 
higher accuracies in the lower frequencies where band power features may be less important \citep{hammer_role_2013}.
Nevertheless, ConvNets failed to clearly outperform FBCSP in our experiments.
Several reasons might contribute to this:
the datasets might still not be large enough to reveal the full
potential of deeper convolutional networks in EEG decoding; or the
class-discriminative features might not have enough hierarchical
structure which deeper ConvNets could exploit. The dataset-size issue
could be solved by either creating larger datasets or also by using
transfer learning approaches across subjects and/or other datasets.
Further analysis of the data itself and of the convolutional networks
might help to shed light whether there are features with a lot of hierarchical
structure. Finally, recurrent networks could exploit signal
changes that happen on longer timescales, e.g., electrodes slowly losing
scalp contact over the course of a session, changes of the electrode cap
position or nonstationarities in the brain signals. Thus, there is
clearly still a large potential for methodological improvement in
ConvNet-based EEG decoding. 

These methodological improvements might also come from further methodological advances in deep learning,
such as newer forms of hyperparameter optimization, in case these advances also translate to even better EEG decoding 
accuracies.
As discussed above, recent advances like dropout, batch normalization and exponential linear units can substantially
improve the performance of EEG decoding with ConvNets, especially for our deep architecture.
Therefore, using other recent techniques, such as newer forms of hyperparameter optimization 
\citep{domhan_speeding_2015,klein_fast_2016,springenberg_bayesian_2016} hold promise to further 
increase accuracies of ConvNets for brain-signal decoding.
Furthermore, as the field is still evolving at a fast pace, new techniques can be expected to be developed
and might then also benefit brain-signal decoders using convolutional neural networks.

However, methodological improvements may also happen in the broad field of ``non-ConvNet'' approaches. 
Obviously, currently no final verdict is possible about an ``optimal'' method for EEG decoding, if
there is a single best method for the large variety of EEG decoding
problems at all. The findings of the present study however support that
ConvNet-based decoding is a contender in this competition.

\subsubsection{Further potential advantages of ConvNets for brain-signal decoding}
Besides the decoding performance, there are also other potential advantages of using deep ConvNets for brain-signal 
decoding.
First, several use cases desirable for brain-signal decoding are very easy to do with deep ConvNets 
iteratively trained in an end-to-end fashion:
Deep ConvNets can be applied to other types of tasks such as as workload estimation, 
error- or event-related potential decoding  (as others have started  \citep{lawhern_eegnet:_2016}) or even other types 
of recordings such as MEG or ECoG.
Also, ConvNets, due to their iterative training, have a natural way of pretraining and finetuning; for example a 
ConvNet can be pretrained on data from the past or data from other subjects and then be finetuned with new data from a 
new subject.
Finetuning can be as simple as continuing the iterative training process on the new data, possibly with a smaller 
learning rate and this finetuning can also be used to perform supervised online adaptation.
Second, due to their joint optimization, single ConvNets can be building blocks for more sophisticated setups of 
multiple ConvNets.
One recent example attempts to create ConvNets that are robust to changes in the input distribution 
\citep{ganin_domain-adversarial_2016}.
This could be used to alleviate the long-standing EEG decoding problem of changes in the EEG signal distribution from 
one session to another.

\subsection{Training strategy}
\subsubsection{Cropped training effect on accuracies}
We observed that cropped training was necessary for the deep ConvNet to reach competitive accuracies on the dataset 
excluding very low frequencies (see Result \ref{subsec:results-cropped}).
The large increase in accuracy with cropped training for the deep network on the 4--$f_{end}$-Hz data might indicate a 
large number of training examples is necessary to learn to extract band power features.
This makes sense as the shifted neighboring windows may contain the same, but shifted, oscillatory signals.
These shifts could prevent the network from overfitting on phase information within the trial, which is less important 
in the higher than the lower frequencies \citep{hammer_role_2013}.
This could also explain why other studies on ConvNets for brain-signal decoding, which did not use cropped training, 
but where band power might be the most discriminative feature, have used fairly shallow architectures and sometimes 
found them to be superior to deeper versions \citep{stober_using_2014}.

\subsubsection{Suitability for online decoding}

Our cropped training strategy appears particularly well-applicable for online brain-signal decoding.
As described above, 
it may offer performance advantages compared with conventional (non-cropped) training.
Additionally, cropped training 
allows for a useful calibration of the trade-off between decoding delay and decoding accuracy in online settings. The 
duration from trial start until the last sample of the first crop should roughly correspond to the minimum time needed 
to decode a control signal. Hence, smaller crops can allow less delay --- the first small crop could end at an early 
sample within the trial without
containing too many timesteps from before the trial that could otherwise disturb the training process. Conversely, 
larger crops that still contain mostly timesteps from within the trial imply a larger delay until a control signal is 
decoded while possibly increasing the decoding accuracy due to more information contained in the larger crops. These 
intuitions should be confirmed in online experiments.

\subsection{Visualization}
\subsubsection{Insights from current visualizations}
In addition to exploring how ConvNets can be successfully used to decode
information from the EEG, we have also developed and tested two
complementary methods to visualize what ConvNets learn from the EEG
data. So far, the literature on using ConvNets for brain-signal decoding
has, for example, visualized weights or outputs of ConvNet layers
\citep{bashivan_learning_2016,
santana_joint_2014,
stober_learning_2016,
yang_use_2015}, determined inputs that maximally activate specific convolutional filters 
\citep{bashivan_learning_2016}, or described attempts at synthesizing the preferred input of a convolutional filter 
\citep{bashivan_learning_2016} (see Supplementary Section  \ref{supp:subsec-related-work} for a more extensive overview).
Here, we applied both a correlative and a causally interpretable visualization method to visualize the frequencies
and spatial distribution of band power features used by the networks.
The visualizations showed plausible spatial distributions for motor
tasks in the alpha, beta and gamma bands (see Section \ref{subsec:results-visualization}). The input-feature unit-output
and the input-perturbation network-prediction correlation maps
together clearly showed that the deep ConvNet learned to extract and use
band power features with specific spatial distributions. Hence, while
the computation of power was built into both the FBCSP and shallow
ConvNet, our deep ConvNets successfully learned to perform the
computation of band power features from the raw input in an end-to-end manner. 
Our network correlation maps can readily show spatial distributions per subject and for the whole group of subjects.
Thus, our network correlation maps proved useful as a technique for spatially mapping the features learned by the ConvNets to perform single-trial decoding.

\subsubsection{Feature discovery through more sophisticated visualizations?}
One limitation of the visualizations presented here is that so far, we only designed them to show how ConvNets use 
known band power features.
However, it could be even more interesting to investigate whether novel or so-far unknown features are used and to 
characterize them. This could be especially informative for tasks where the discriminative features are less well known 
than for motor decoding, e.g. for less-investigated tasks such as decoding of task performance 
\citep{meinel_pre-trial_2016}.
But even for the data used in this study, our results show hints that deep ConvNets used different 
features than shallow ConvNets as well as the FBCSP-based decoding, since there are statistically significant 
differences between their confusion matrices (see Result \ref{subsec:results-confmats}).
This further strengthens the motivation to explore what features the deep ConvNet exploits, 
for example using visualizations that show what parts of a trial are relevant for the classification decision or what a 
specific convolutional filter/unit output encodes.
Newer visualization methods such as layer-wise relevance propagation 
\citep{bach_pixel-wise_2015,montavon_explaining_2017}, 
inverting convolutional networks with convolutional networks \citep{dosovitskiy_inverting_2016} or synthesizing 
preferred inputs of units \citep{nguyen_synthesizing_2016} could be promising next steps in that direction.

\subsection{Conclusion}
In conclusion, ConvNets are not only a novel, promising tool in the EEG decoding toolbox,
but combined with innovative visualization techniques, they may also open up new windows for EEG-based brain mapping.

\section*{Acknowledgements}
This work was supported by the BrainLinks-BrainTools Cluster of Excellence (DFG grant EXC1086) and by the Federal 
Ministry of Education and Research (BMBF, grant Motor-BIC 13GW0053D).

\subsection*{Conflicts of interest}
The authors declare that there is no conflict of interest regarding the publication of this paper.

\bibliographystyle{apalike}
\bibliography{bibliography}

\appendix
\beginsupplement

\newcommand{\cellbr}[0]{\vspace{0.1cm} \hspace{10cm}} 
\newgeometry{margin=2cm} 
\begin{landscape}
\section{Supplementary Materials}
\subsection{Related Work}
\label{supp:subsec-related-work}
\singlespacing
\scriptsize
\renewcommand*{\arraystretch}{2.0}
\begin{longtable}{p{0.2\textwidth}p{0.11\textwidth}p{0.06\textwidth}p{0.1\textwidth}
p{0.13\textwidth}p{0.13\textwidth}p{0.1\textwidth}p{0.11\textwidth}p{0.2\textwidth}}
\toprule
\textbf{\small Study} &
\textbf{\small Decoding problem} &
\textbf{\small Input domain} &
\textbf{\small Conv/ dense layers} &
\textbf{\small Design choices} &
\textbf{\small Training strategies} &
\textbf{\small External baseline} &
\textbf{\small Visualization type(s)} &
\textbf{\small Visualization findings} \\
\midrule
\endhead

\hdashline

This manuscript, Schirrmeister et. al (2017) &
Imagined and executed movement classes, within subject &
Time, \hspace{1cm} 0--125 Hz & 5/1 &
Different ConvNet architectures \cellbr
Nonlinearities and pooling modes \cellbr
Regularization and intermediate normalization layers \cellbr
Factorized convolutions \cellbr
Splitted vs one-step convolutions &
Trial-wise vs. cropped training strategy &
FBCSP + rLDA & 
Feature activation correlation \cellbr
Feature-perturbation prediction correlation &
See Section \ref{subsec:results-visualization}
\\
\hdashline 

Single-trial EEG classification of motor imagery using deep convolutional neural networks, \citet{tang_single-trial_2017} &
Imagined movement classes, within-subject &
Time, \hspace{1cm} 8--30 Hz & 2/2 & & 
& & &
\\
\hdashline 
 
EEGNet: A Compact Convolutional Network for EEG-based Brain-Computer Interfaces, \citet{lawhern_eegnet:_2016} & 
Oddball response (RSVP), error response (ERN), movement classes (voluntarily started and imagined) &
Time, 0.1--40 Hz & 3/1 &  Kernel sizes & 
&   & &
\\
\hdashline
 
Remembered or Forgotten? --- An EEG-Based Computational Prediction Approach, \citet{sun_remembered_2016} & 
Memory performance, within-subject &
Time, 0.05--15 Hz & 2/2 & & Different time windows &  &
Weights (spatial) & 
Largest weights found over prefrontal and temporal cortex
\\
\hdashline

Multimodal Neural Network for Rapid Serial Visual Presentation Brain Computer Interface, \citet{manor_multimodal_2016}
&
Oddball response using RSVP and image (combined image-EEG decoding), within-subject&
Time, 0.3--20 Hz & 3/2 & & &  & 
Weights \cellbr Activations \cellbr Saliency maps by gradient &
Weights showed typical P300 distribution \cellbr
Activations were high at plausible times (300-500ms) \cellbr
Saliency maps showed plausible spatio-temporal plots
\\
\hdashline
 
A novel deep learning approach for classification of EEG motor imagery signals, \citet{tabar_novel_2017} &
Imagined and executed movement classes, within-subject &
Frequency, 6--30 Hz & 1/1 & 
\multicolumn{2}{p{0.285\textwidth}}{Addition of six-layer stacked autoencoder on ConvNet features \cellbr Kernel sizes} 
& FBCSP, Twin SVM, DDFBS, Bi-spectrum, RQNN  & Weights (spatial + frequential) &
Some weights represented difference of values of two electrodes on different sides of head
\\
\hdashline
 
Predicting Seizures from Electroencephalography Recordings: A Knowledge Transfer Strategy, \citet{liang_predicting_2016} &
Seizure prediction, within-subject & Frequency, 0--200 Hz & 1/2 & & 
Different subdivisions of frequency range \cellbr
Different lengths of time crops \cellbr
Transfer learning with auxiliary non-epilepsy datasets &
& Weights \cellbr Clustering of weights &
Clusters of weights showed typical frequency band subdivision (delta, theta, alpha, beta, gamma)
\\
\hdashline
 
EEG-based prediction of driver's cognitive performance by deep convolutional neural network, \citet{hajinoroozi_eeg-based_2016} &
Driver performance, within- and cross-subject &
Time, \hspace{1cm} 1--50 Hz & 1/3 &
\multicolumn{2}{p{0.285\textwidth}}{Replacement of convolutional layers by restricted Boltzmann machines with slightly varied network architecture}  & 
&
\\
\hdashline
 
Deep learning for epileptic intracranial EEG data, \citet{antoniades_deep_2016} &
Epileptic discharges, cross-subject & Time, \hspace{1cm} 0--100 HZ & 1--2/2 & 1 or 2 convolutional layers &  & &
Weights \cellbr
Correlation weights and interictal epileptic discharges (IED) \cellbr
Activations &
Weights increasingly correlated with IED waveforms with increasing number of training iterations \cellbr
Second layer captured more complex and well-defined epileptic shapes than first layer \cellbr
IEDs led to highly synchronized activations for neighbouring electrodes
\\
\hdashline
 
Learning Robust Features using Deep Learning for Automatic Seizure Detection, \citet{thodoroff_learning_2016} &
Start of epileptic seizure, within- and cross-subject &
Frequency, mean amplitude for 0--7 Hz, 7--14 Hz, 14--49 Hz & 3/1 (+ LSTM as postprocessor) & &
 & Hand crafted features + SVM & Input occlusion and effect on prediction accuracy &
Allowed to locate areas critical for seizure 
\\
\hdashline

Single-trial EEG RSVP classification using convolutional neural networks, \citet{george_single-trial_2016} &
Oddball response (RSVP), groupwise (ConvNet trained on all subjects) &
Time, 0.5--50 Hz & 4/3 & &  &  &
Weights (spatial) &
Some filter weights had expected topographic distributions for P300 \cellbr
Others filters had large weights on areas not traditionally associated with P300
\\
\hdashline

Wearable seizure detection using convolutional neural networks with transfer learning, \citet{page_wearable_2016} &
Seizure detection, cross-subject, within-subject, groupwise &
Time, \hspace{1cm} 0--128 Hz & 1-3/1-3 & & Cross-subject supervised training, within-subject finetuning of fully connected layers &
Multiple: spectral features, higher order statistics + linear-SVM, RBF-SVM, ...& & 
\\
\hdashline

Learning Representations from EEG with Deep Recurrent-Convolutional Neural Networks, \citet{bashivan_learning_2016}  &
Cognitive load (number of characters to memorize), cross-subject & 
Frequency, mean power for 4--7 Hz, 8--13 Hz, 13--30 Hz & 3--7/2 (+ LSTM or other temporal post-processing (see design choices)) &
Number of convolutional layers \cellbr
Temporal processing of ConvNet output by max pooling, temporal convolution, LSTM or temporal convolution + LSTM & & &
Inputs that maximally activate given filter \cellbr
Activations of these inputs \cellbr
"Deconvolution" for these inputs &
Different filters were sensitive to different frequency bands \cellbr
Later layers had more spatially localized activations \cellbr
Learned features had noticeable links to well-known electrophysiological markers of cognitive load \cellbr
\\
\hdashline

Deep Feature Learning for EEG Recordings, \citet{stober_learning_2016} &
Type of music rhythm, groupwise (ensembles of leave-one-subject-out trained models, evaluated on separate test set of same subjects) &
Time, 0.5--30Hz & 2/1 & Kernel sizes & 
Pretraining first layer as convolutional autoencoder with different constraints &  & 
Weights (spatial+3 timesteps, pretrained as autoencoder) & 
Different constraints led to different weights, one type of constraints could enforce weights that are similar across subjects; other type of constraints led to weights that have similar spatial topographies under different architectural configurations and preprocessings
\\
\hdashline

Convolutional Neural Network for Multi-Category Rapid Serial Visual Presentation BCI, \citet{manor_convolutional_2015} &
Oddball response (RSVP), within-subject &
Time, 0.1--50 Hz & 3/3 (Spatio-temporal regularization) && &&
Weights \cellbr Mean and single-trial activations &
Spatiotemporal regularization led to softer peaks in weights \cellbr
Spatial weights showed typical P300 distribution \cellbr
Activations mostly had peaks at typical times (300-400ms)
\\
\hdashline

Parallel Convolutional-Linear Neural Network for Motor Imagery Classification, \citet{sakhavi_parallel_2015}  &
Imagined movement classes, within-subject &
Frequency, 4--40 Hz, using FBCSP & 2/2 (Final fully connected layer uses concatenated output by convolutional
and fully connected layers) &
Combination ConvNet and MLP (trained on different features) vs. only ConvNet vs. only MLP & 
& & & 
\\
\hdashline

Using Convolutional Neural networks to Recognize Rhythm Stimuli form Electroencephalography Recordings, \citet{stober_using_2014} &
Type of music rhythm, within-subject & Time and frequency evaluated, 0-200 Hz & 1-2/1 &
Best values from automatic hyperparameter optimization: frequency cutoff, one vs two layers, kernel sizes, number of channels, pooling width &
Best values from automatic hyperparameter optimization: learning rate, learning rate decay, momentum, final momentum &
&
\\
\hdashline

Convolutional deep belief networks for feature extraction of EEG signal, \citet{ren_convolutional_2014}  &
Imagined movement classes, within-subject &
Frequency, 8--30 Hz &
2/0 (Convolutional deep belief network, separately trained RBF-SVM classifier) & & 
\\
\hdashline

Deep feature learning using target priors with applications in ECoG signal decoding for BCI, \citet{wang_deep_2013}  &
Finger flexion trajectory (regression), within-subject &
Time, 0.15--200 Hz & 3/1 (Convolutional layers trained as convolutional stacked autoencoder with target prior) &
Partially supervised CSA & 
\\
\hdashline

Convolutional neural networks for P300 detection with application to brain-computer interfaces, \citet{cecotti_convolutional_2011}  &
Oddball / attention response using P300 speller, within-subject & Time, 0.1-20 Hz & 2/2 &
Electrode subset (fixed or automatically determined) \cellbr
Using only one spatial filter \cellbr
Different ensembling strategies & 
&
Multiple: Linear SVM, gradient boosting, E-SVM, S-SVM, mLVQ, LDA, ... &
Weights &
Spatial filters were similar for different architectures \cellbr
Spatial filters were different (more focal, more diffuse) for different subjects
\\
\hdashline

\caption{\small{\textbf{Related previous publications using convolutional neural networks for EEG decoding.}
Frequency domain input always only contains amplitudes or a transformation of amplitudes (power, log power, etc.), never phase information.
The number of dense layers includes parametrized classification layers. Layer numbers always refer to EEG decoding models in cases of articles that use multiple modalities for decoding.
Special features of the model written in parentheses after the number of layers, especially if these features make the number of layers misleading.
External baseline: the study includes directly comparable baseline accuracies of non-deep-learning approaches of other authors.
Visualization types and findings both refer to visualizations of the trained networks used for EEG decoding; findings are paraphrased from the original publications.
Note that none of the previous studies using time-domain input showed, using a suitable visualization technique, that the ConvNets learned to use band power features, in contrast to our present study.
Also note that other previous studies used artificial neural networks without convolutions for EEG analysis, e.g. \citet{santana_joint_2014, sturm_interpretable_2016}.}}
\end{longtable}
\end{landscape}
\restoregeometry

\subsection{FBCSP implementation}\label{subsec:supp-fbcsp}

As in many previous studies \citep{lotte_review_2007}, we used regularized linear discriminant analysis (RLDA)
as the classifier, with shrinkage regularization \citep{ledoit_well-conditioned_2004}.
To decode multiple classes, we used one-vs-one majority weighted voting:
We trained an RLDA classifier for each pair of classes, summed the classifier outputs (scaled to be in the same range) 
across classes and picked the class with the highest sum \citep{chin_multi-class_2009,galar_overview_2011}.

FBCSP is typically used with feature selection, since few spatial
filters from few frequency bands often suffice to reach good accuracies
and using many or even all spatial filters often leads to overfitting
\citep{blankertz_optimizing_2008,chin_multi-class_2009}.
We use a classical measure for preselecting spatial
filters, the ratio of the corresponding power features for both classes extracted by each spatial filter
\citep{blankertz_optimizing_2008}. 
Additionally, we performed a feature selection step on
the final filter bank features by selecting features using an inner
cross validation on the training set, see published code 
\footnote{\url{
https://github.com/robintibor/braindecode/blob/f9497f96a6dfdea1e24a4709a9ceb30e0f4768e3/braindecode/csp/pipeline.py\#L20
0-L408}} for
details.

In the present study, we designed two filter banks adapted for the two datasets to capture most 
discriminative motor-related band power information.
In preliminary experiments, overlapping frequency bands led to higher accuracies, as also proposed by
\citet{sun_novel_2010}.
As the bandwidth of physiological EEG power modulations typically increases in higher frequency ranges 
\citep{buzsaki_neuronal_2004},
we used frequency bands with 6 Hz width and 3 Hz overlap in frequencies up to 13 Hz, and bands of 8 Hz width and 4 Hz 
overlap in the range above 10 Hz.
Frequencies above 38 Hz only improved accuracies on one of our datasets, the so-called High-Gamma Dataset
(see Section \ref{subsec:datasets}, where we also describe the likely reason for this difference, namely that the 
recording procedure for the High-Gamma Dataset --- 
but not for the BCI Competition Dataset --- was specifically optimized for the high frequency range).
Hence the upper limit of used frequencies was set at 38 Hz for the BCI Competition Dataset, while the upper limit for 
the High-Gamma Dataset was set to 122 Hz, close to the Nyquist frequency, thus allowing FBCSP to also use 
information from the gamma band.

As a sanity check, we compared the accuracies of our FBCSP implementation to those published in the literature for the 
same BCI Competition Dataset \citep{sakhavi_parallel_2015}, showing very similar performance:
67.59\% for our implementation vs 67.01\% for their implementation on average across subjects (p>0.7, Wilcoxon 
signed-rank test, see Result \ref{subsec:results-baseline} for more detailed results).
This underlines that our FBCSP implementation, including our feature selection and filter bank design, indeed was a 
suitable baseline for the evaluation of our ConvNet
decoding accuracies.

\subsection{Residual network architecture}\label{subsec:supp-resnet-architecture}
In total, the ResNet has 31 convolutional layers, a depth where ConvNets without residual blocks started to show 
problems converging in the original ResNet paper \citep{he_deep_2015}. In layers where the number of channels is 
increased, we padded the incoming feature map with zeros to 
match the new channel dimensionality for the shortcut,
as in option A of the original paper \citep{he_deep_2015}.

\begin{table}[H]
\centering \footnotesize
\begin{tabular}{llll}\toprule
\textbf{\normalsize Layer/Block} & \textbf{\normalsize Number of Kernels} & \textbf{\normalsize Kernel Size} & 
\textbf{\normalsize Output Size} \\\midrule
 Input &  &  & 1000x44x1 \\ 
 Convolution (linear) & 48 & 3x1 & 1000x44x48 \\
 Convolution (ELU) & 48 & 1x44 & 1000x1x48 \\
 ResBlock (ELU) & 48 & 3x1 & \\
 ResBlock (ELU) & 48 & 3x1 & \\
 ResBlock (ELU) & 96 & 3x1 (Stride 2x1) & 500x1x96\\
 ResBlock (ELU) & 96 & 3x1  & \\
 ResBlock (ELU) & 144 & 3x1 (Stride 2x1) & 250x1x96\\
 ResBlock (ELU) & 144 & 3x1  & \\
 ResBlock (ELU) & 144 & 3x1 (Stride 2x1) & 125x1x96\\
 ResBlock (ELU) & 144 & 3x1  & \\
 ResBlock (ELU) & 144 & 3x1 (Stride 2x1) & 63x1x96\\
 ResBlock (ELU) & 144 & 3x1  & \\
 ResBlock (ELU) & 144 & 3x1 (Stride 2x1) & 32x1x96\\
 ResBlock (ELU) & 144 & 3x1  & \\
 ResBlock (ELU) & 144 & 3x1 (Stride 2x1) & 16x1x96\\
 ResBlock (ELU) & 144 & 3x1  & \\
 Mean Pooling & & 10x1& 7x1x144 \\
 Convolution + Softmax & 4 & 1x1 & 7x1x4 \\
 \\ \bottomrule
\hline
\end{tabular}
\caption{\textbf{Residual network architecture hyperparameters}.
Number of kernels, kernel and output size for all subparts of the network.
Output size is always time x height x channels.
Note that channels here refers to input channels of a network layer, not to EEG channels;
EEG channels are in the height dimension.
Output size is only shown if it changes from the previous block.
Second convolution and all residual blocks used ELU nonlinearities.
Note that in the end we had seven outputs, i.e., predictions for the four classes, in the time dimension ( 
\textbf{7}x1x4 final output size).
In practice, when using cropped training as explained in Section \ref{subsec:cropped-training}, we even had 424 
predictions, and used the mean of these to predict the trial.}
\label{table:resnet-hyperparams}
\end{table}

\subsection{Optimization and early stopping}\label{subsec:supp-optimization}
Adam is a variant of stochastic gradient descent designed to work well with high-dimensional parameters, which makes it 
suitable for optimizing the large number of parameters of a ConvNet \citep{kingma_adam:_2014}.
The early stopping strategy that we use throughout this study, developed in the computer vision field
\footnote{\url{
https://web.archive.org/web/20160809230156/https://code.google.com/p/cuda-convnet/wiki/Methodology}}, 
splits the training set into a training and validation fold and stops the first phase of the training when validation 
accuracy does not improve for a predefined number of epochs.
The training continues on the combined training and validation fold starting from the parameter values that led to the 
best
accuracies on the validation fold so far.
The training ends when the loss function on the validation fold drops to the same value as the loss function on the 
training fold at the end of the first training phase (we do not continue training in a
third phase as in the original description).
Early stopping in general allows training on different types of networks and datasets without choosing the number of 
training epochs
by hand. Our specific strategy uses the entire training data while only training once.
In our study, all reported accuracies have been determined on an independent test set.

\subsection{Visualization methods}\label{subsec:supp-visualization-methods}

\subsubsection{Input-feature unit-output correlation maps}\label{subsec:correlation-visualization}

\begin{figure}
\begin{subfigure}[t]{\linewidth}
\centering

\includegraphics[max size={1\linewidth}{0.2\paperheight}]{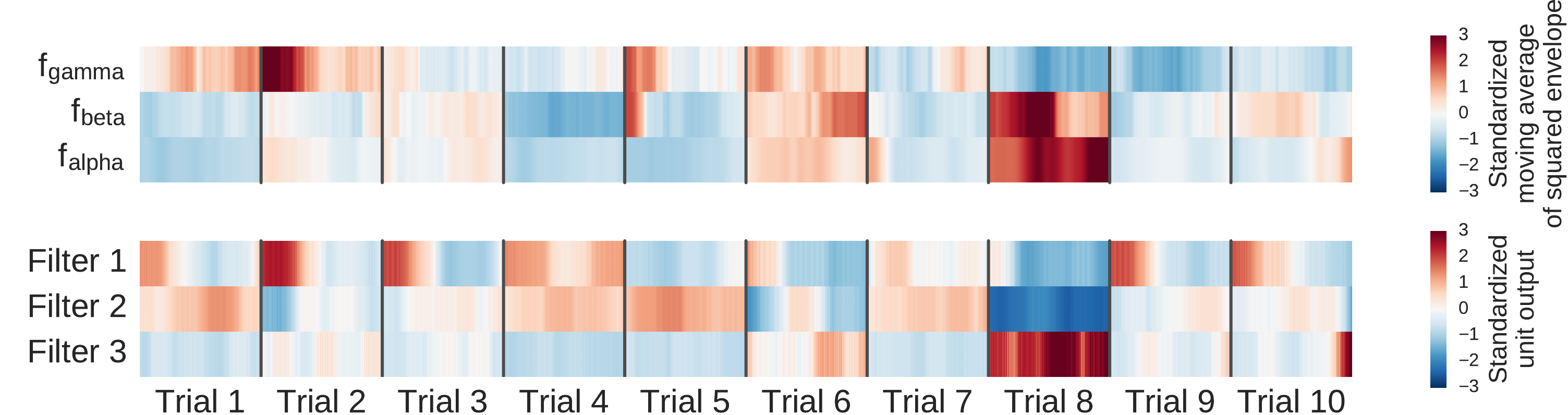}
\caption{\scriptsize{Feature inputs and unit outputs for input-feature unit-output
correlation map. Moving average of squared envelopes and unit outputs
for 10 trials. Upper rows show mean squared envelopes over the receptive
field for three frequency ranges in the alpha, beta and gamma frequency
band, standardized per frequency range. Lower rows show corresponding
unit outputs for three filters, standardized per filter. All time series
standardized for the visualization.}}
\label{fig:computation-correlation-a}
\end{subfigure}%
\\
\begin{subfigure}[t]{\linewidth}
\centering
\includegraphics[max size={0.7\linewidth}{0.2\paperheight}]{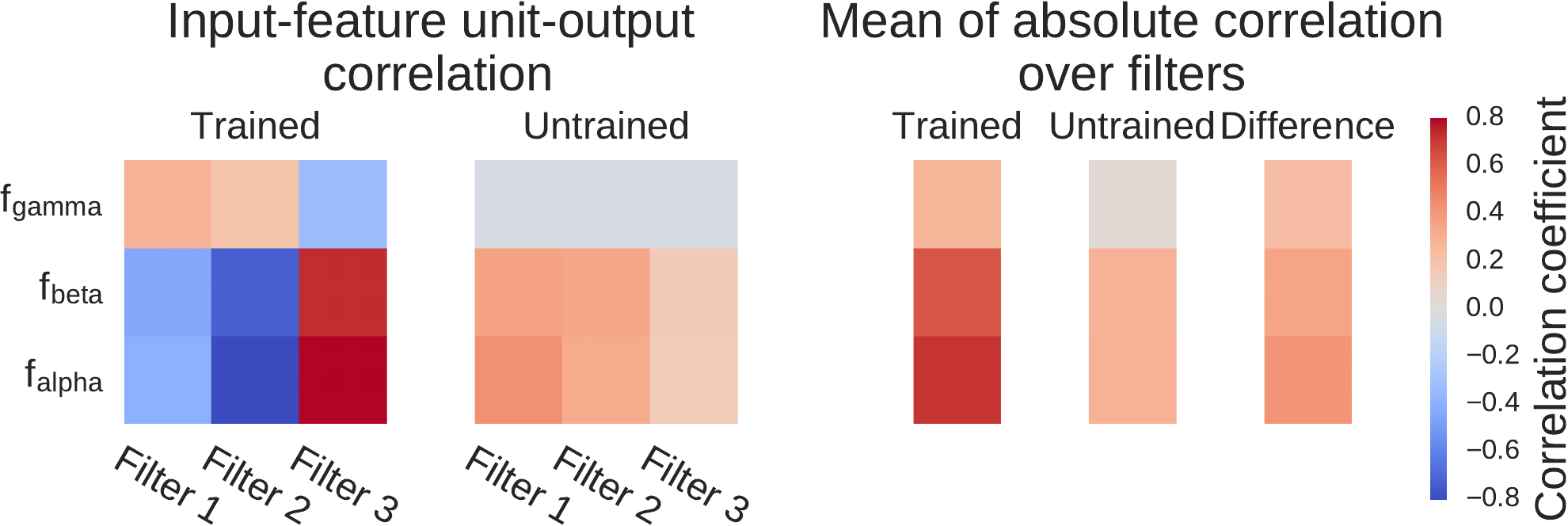}
\includegraphics[max size={0.27\linewidth}{0.2\paperheight}]{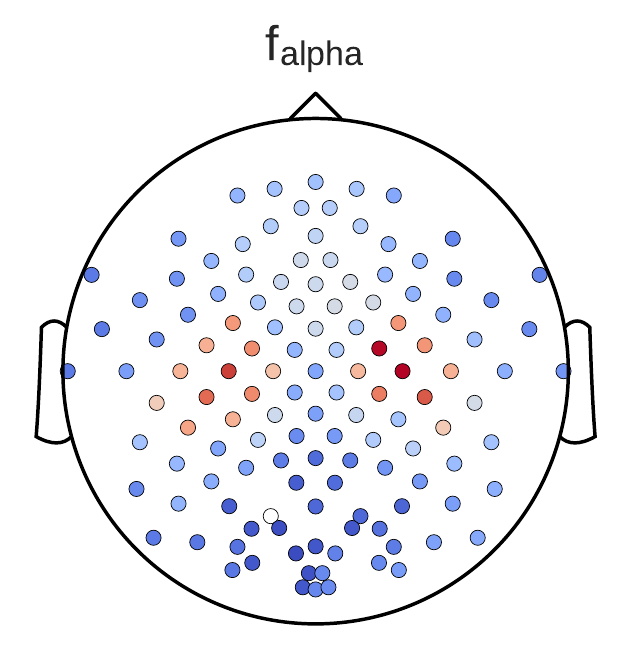}
\caption{\scriptsize{Input-feature unit-output correlations and corresponding scalp map for the alpha band.
Left: Correlation coefficients between unit outputs of three filters and mean squared envelope values over the
corresponding receptive field of the units for three frequency ranges in the alpha (7--13 Hz),
beta (13--31 Hz), and gamma (71--91 Hz) frequency band.
Results are shown for the trained and the untrained ConvNet and for one electrode.
Middle: Mean of the absolute correlation coefficients over the three filters for the trained and the
untrained ConvNet, and the difference between trained and untrained ConvNet.
Right: An exemplary scalp map for correlations in the alpha band (7--13 Hz), where the color of each dot encodes 
the correlation difference between a trained and an untrained ConvNet for one electrode.
Note localized positive effects above areas corresponding to the right and left sensorimotor hand/arm areas,
indicating that activity in these areas has large absolute correlations with the predictions of the trained ConvNet.}}
\label{fig:computation-correlation-b}
\end{subfigure}
\caption{\textbf{
Computation overview for input-feature unit-output network correlation map.}}
\label{fig:computation-correlation}
\end{figure}

\begin{figure}
\centering
\includegraphics[max size={1\linewidth}{0.2\paperheight}]{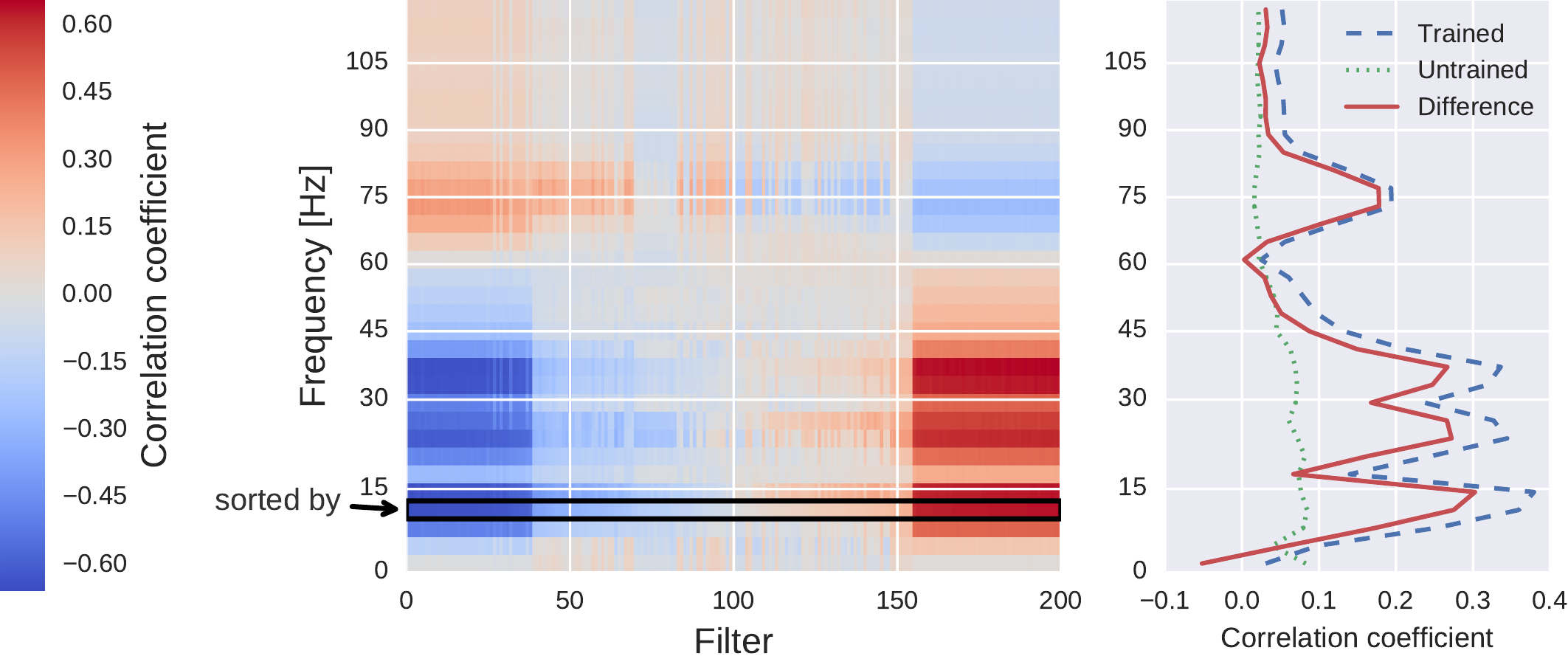}
\caption{\textbf{Correlation between the mean squared envelope feature and unit output
for a single subject at one electrode position (FCC4h).}
Left: All correlations.
Colors indicate the correlation between unit outputs per convolutional
  filter (x-axis) and mean squared envelope in different frequency bands
  (y-axis). Filters are sorted by their correlation to the 7--13 Hz
  envelope (outlined by the black rectangle). Note the large
  correlations/anti-correlations in the alpha/beta bands (7--31 Hz)
  and somewhat weaker correlations/anti-correlations in the gamma band
  (around 75 Hz).
  Right: Mean absolute values across units of all convolutional filters for all
  correlation coefficients of the trained model, the untrained model and
  the difference between the trained and untrained model. Peaks in the
  alpha, beta and gamma bands are clearly visible.}
\label{fig:correlation-single-subject-electrode}
\end{figure}%

The input-feature unit-output correlation maps visualize the frequency-resolved correlation between unit outputs
of the convolutional filters of the ConvNets and the power of all the samples in the receptive field of these units 
(see Figure \ref{fig:computation-correlation}).

To achieve this,we performed the following steps:

\begin{enumerate}
 \item For each frequency band of interest, the signal was bandpass-filtered to that frequency band and the envelope 
was computed.
\item For each frequency band of interest, the squared mean envelope for each receptive field of a 
given layer was computed. We did this by computing a moving window 
average of the squared envelope with the moving window size the same as the receptive field size (this was the input 
feature for which we then evaluated how much it affected the unit output).
\item Unit outputs of the given layer on the original signal were computed.
\item Linear correlations between the squared mean envelope values for all the frequency bands and the 
unit outputs for each convolutional filter were computed. These correlations should reflect whether a filter might 
compute an approximation of the squared mean envelope of all the samples in its receptive field.
\end{enumerate}
Since we compute correlations after concatenating all samples of all trials, 
these correlations reflect both within-trial and between-trial effects.
The proposed method could, however, be straightforwardly extended to disentangle these two sources.
We computed the correlations for a filter bank ranging from 0 to 119 Hz.
An example result for a single electrode and a single subject is shown in Figure 
\ref{fig:correlation-single-subject-electrode}.

To compute a single scalp plot for a frequency band, we computed the
mean of the absolute correlations over all units for each convolutional
filter and each electrode for that frequency band.
To disentangle effects which are caused by the training of the network from those caused by the architecture,
we computed the scalp plot for a trained and an untrained model.
The scalp plot for a subject is then the scalp plot of the trained model minus the scalp plot of the untrained model 
(see Figure \ref{fig:computation-correlation-b}).
The group scalp plot is the mean of these differential scalp plots over all subjects.

To compare the resulting maps against scalp maps that simply result from class-feature correlations,
we also computed the linear correlation between mean squared envelope values and the one-hot-encoded classes, in the 
same way as before.
First, for each trial, each sensor and each frequency band, we constructed a vector of the moving window 
squared envelope values as before
(with the moving window now the size of the receptive field of the last layer of the ConvNet).
Second, for each trial and each class, we constructed a vector of either value 1 if the trial was of the given class
or value 0 if it was of another class. The concatenated vectors then resulted in a time series with value 1 if the 
time point belonged to a given class and value 0 if it did not.
Then we correlated the moving window squared envelope time series vectors with the class time series vectors,
resulting in one correlation value per class, sensor and frequency band combination.
As in the other computations, we subtracted the correlations resulting from
predictions of an untrained deep ConvNet. 

A further question is whether the correlations could be a result of the unit outputs encoding the 
final class label. Such correlations could also result from using other discriminative features than the features we 
analyzed.
To investigate this question, we correlated the unit outputs for each layer with the class labels.
Here, we proceeded the same way as described in the previous paragraph, but correlated unit outputs directly with 
class labels.
We then computed a single absolute correlation coefficient per layer in two ways:
First, we computed the mean absolute correlation coefficient for all classes and all filters. These correlations 
should show how strongly the unit outputs encode the class labels on average across filters.
Second, we computed the maximum absolute correlation coefficients for each class over all filters and then computed
the mean of these maxima of the four classes. 
These correlations should show how strongly the unit outputs encode the class labels for the most 
``class-informative'' filters.
Finally, for both versions and as in the other visualizations, we also computed the difference of these 
correlations between a trained and an untrained model. In summary, this approach allowed to show how unit-output class 
correlations arise from layer to layer through the ConvNet.

\subsubsection{Input-perturbation network-prediction correlation
map}\label{subsec:perturbation-visualization}

\begin{figure}
\begin{subfigure}[t]{\linewidth}
\centering
\includegraphics[max size={1\linewidth}{0.3\paperheight}]{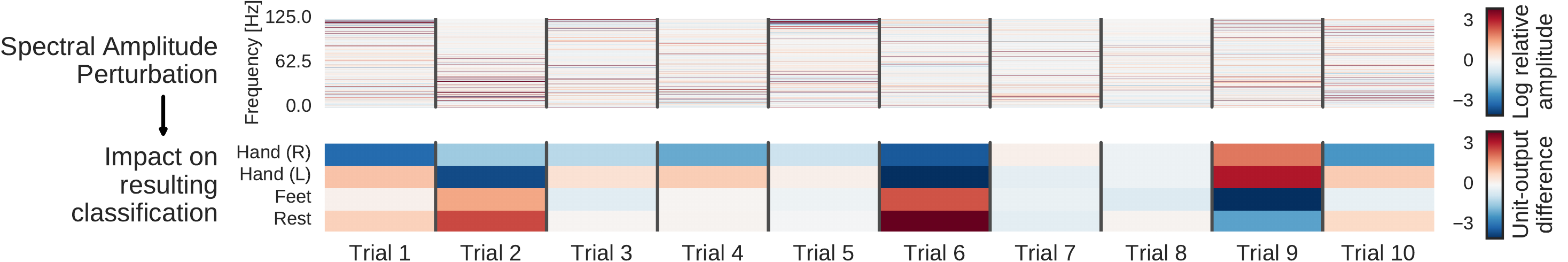}
\caption{\scriptsize{Example spectral amplitude perturbation and resulting classification
difference. Top: Spectral amplitude perturbation as used to perturb the
trials. Bottom: unit-output difference between unperturbed and perturbed
trials for the classification layer units before the softmax.}}
\label{fig:computation-perturbation-a}
\end{subfigure}
\\
\begin{subfigure}[t]{\linewidth}
\centering
\includegraphics[max size={0.31\linewidth}{0.2\paperheight}]{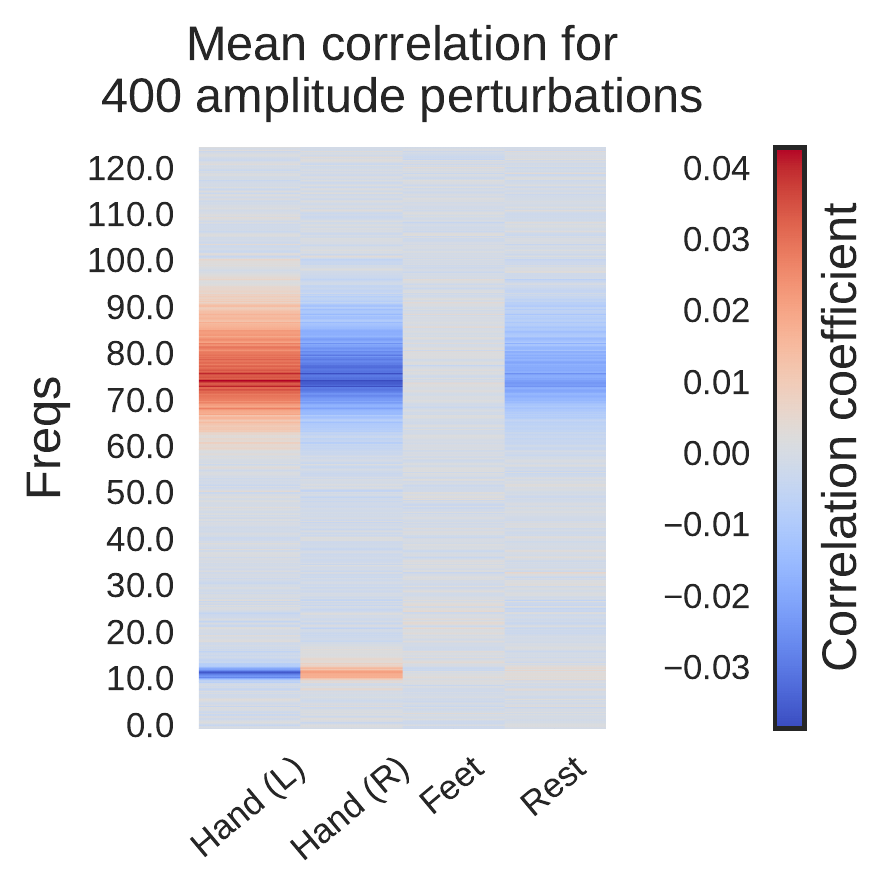}
\includegraphics[max size={0.31\linewidth}{0.2\paperheight}]{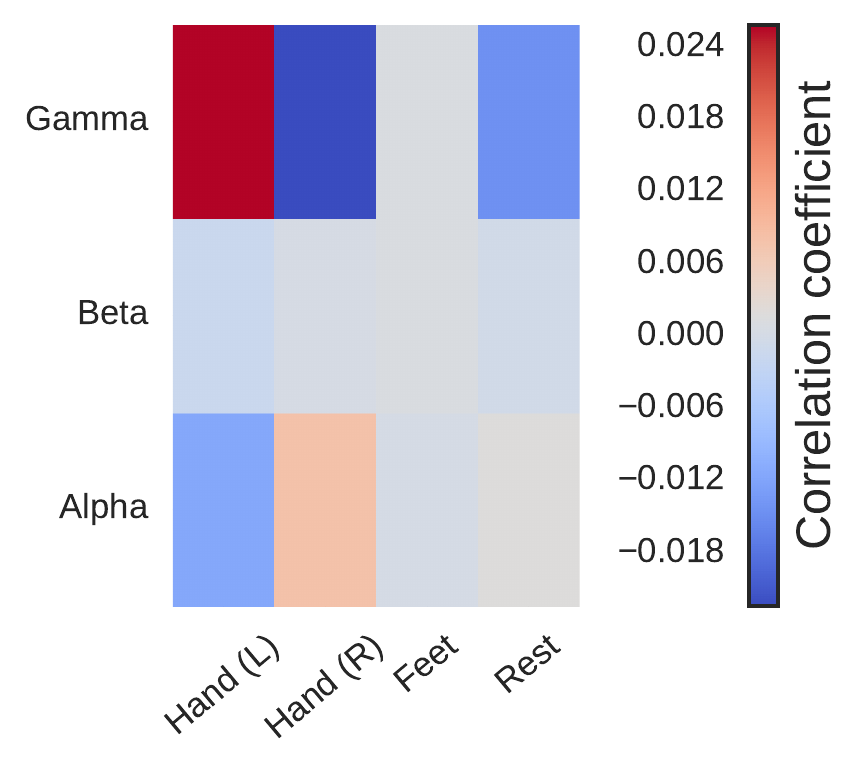}
\includegraphics[max size={0.32\linewidth}{0.2\paperheight}]{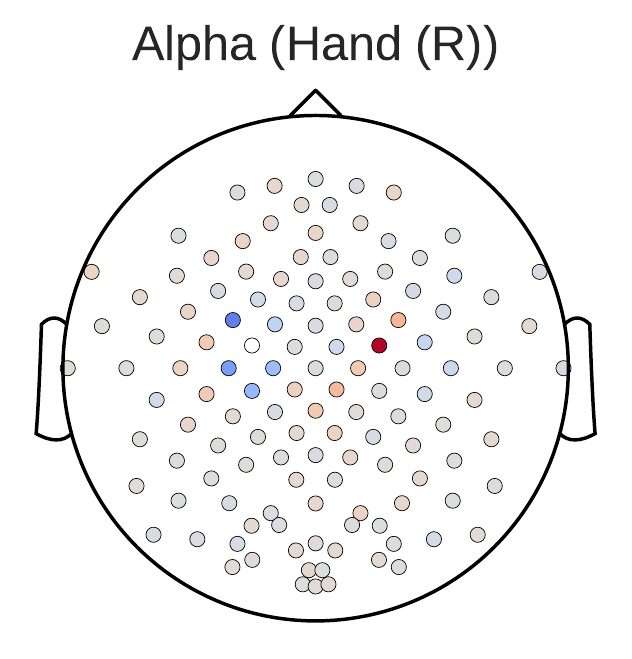}
\caption{\scriptsize{Input-perturbation network-prediction correlations and corresponding
network correlation scalp map for alpha band.
Left: Correlation
coefficients between spectral amplitude perturbations for all frequency
bins and differences of the unit outputs for the four classes
(differences between unperturbed and perturbed trials) for one
electrode.
Middle: Mean of the correlation coefficients over the the
alpha (7--13 Hz), beta (13--31 Hz) and gamma (71--91 Hz) frequency ranges.
Right: An exemplary scalp map for the alpha band, where the color of each dot encodes
the correlation of amplitude changes at that electrode and the corresponding prediction changes of the ConvNet.
Negative correlations on the left sensorimotor hand/arm areas show an amplitude decrease in these areas
leads to a prediction increase for the Hand (R) class, whereas positive correlations on the right sensorimotor hand/arm 
areas
show an amplitude decrease leads to a prediction decrease for the Hand (R) class.
This complements the information from the input-feature unit-output network correlation map
(see Figure \ref{fig:computation-correlation-b}), which 
showed band power in these areas is strongly correlated with unit outputs in the penultimate layer.
}}
\label{fig:computation-perturbation-b}
\end{subfigure}
\caption{\textbf{
Computation overview for input-perturbation network-prediction correlation map.}}
\label{fig:computation-perturbation}
\end{figure}

To investigate the causal effect of changes in power on the deep ConvNet,
we correlated changes in ConvNet predictions with changes in amplitudes by perturbing 
the original trial amplitudes (see figure \ref{fig:computation-perturbation} for an overview).
Concretely, we transformed all training trials into the frequency domain by a Fourier transformation. 
Then we randomly perturbed the amplitudes by adding Gaussian noise (with mean 0 and variance 1) to them. 
The phases were kept unperturbed. 
After the perturbation, we re-transformed to the time domain by the inverse Fourier transformation. 
We computed predictions of the deep ConvNet for these trials before and after the perturbation
(predictions here refers to outputs of the ConvNet directly before the softmax activation). 
We repeated this procedure with 400 perturbations sampled from aforementioned Gaussian distribution 
and then correlated the change in input amplitudes (i.e., the perturbation/noise we added) with the change in the 
ConvNet predictions. 
To ensure that the effects of our perturbations reflect the behavior of the ConvNet on realistic data, we also checked
that the perturbed input does not cause the ConvNet to misclassify the trials  (as can easily happen even from small 
perturbations, see \citet{szegedy_intriguing_2014}).
For that, we computed accuracies on the perturbed trials.
For all perturbations of the training sets of all subjects, accuracies stayed above 99.5\% of the accuracies achieved 
with the unperturbed data.

\subsubsection{EEG spectral power topographies}\label{subsec:spectral-power-topo}

To visualize the class-specific EEG spectral power modulations, we
computed band-specific envelope-class correlations in the alpha, beta
and gamma bands for all classes of the High-Gamma Dataset. The
group-averaged topographies of these maps could be readily compared to
our input-feature unit-output network correlation maps, since, similar
to the power-class correlation map described in Section
\ref{subsec:correlation-visualization}, we computed
correlations of the moving average of the squared envelope with the
actual class labels, using the receptive field size of the final layer as
the moving average window size.
Since this is a ConvNet-independent
visualization, we did not subtract any values of an untrained ConvNet.
We show the resulting scalp maps for the four classes and did not
average over them. Note that these computations were only used for the
power topographies shown in Figure \ref{fig:results-spectral-topo} and did not enter the decoding
analyses as described in the preceding sections.

\subsection{Dataset details}\label{subsec:supp-datasets}

The BCI Competition IV dataset 2a (from here on referred to as BCI Competition Dataset)
is a 22-electrode EEG motor-imagery dataset, with 9 subjects and 2 sessions, each with 288 four-second trials of 
imagined movements per subject (movements of the left hand, the right hand, the feet and the tongue) 
\citep{brunner_bci_2008}.
The training set consists of the 288 trials of the first session, the test set of the 288 trials of the second session.

Our ``High-Gamma Dataset'' is a 128-electrode dataset (of which we later only use 44 sensors 
covering the motor cortex, (see Section \ref{subsec:preprocessing}),
obtained from 20 healthy subjects (9 female, 4 lefthanded, age 27.5$\pm$3.2 (mean$\pm$std)) with
roughly 1000 (992$\pm$135.3, mean$\pm$std) four-second trials of executed movements divided into 13 runs per subject. 
The four classes of movements were movements of either the left hand, the right hand, both feet,
and rest (no movement, but same type of visual cue as for the other classes).
The training set consists of the approx. 880 trials of all runs except the last two runs, the test set of the approx. 
160 trials of the last 2 runs.
This dataset was acquired in an EEG lab optimized for non-invasive detection of high-frequency movement-related EEG 
components \citep{ball_movement_2008,darvas_high_2010}.
Such high-frequency components in the range of approx. 60 to above 100 Hz are typically increased during movement 
execution and may contain useful movement-related 
information \citep{crone_functional_1998,hammer_predominance_2016,quandt_single_2012}.
Our technical EEG Setup comprised
(1.) Active electromagnetic shielding: optimized for frequencies from DC - 10 kHz (-30 dB to -50 dB), shielded window, 
ventilation \& cable
feedthrough (mrShield, CFW EMV-Consulting AG, Reute, CH) (2.) Suitable
amplifiers: high-resolution (24 bits/sample) and low-noise
(\textless{}0.6  $\mu V$ RMS 0.16--200 Hz, \textless{}1.5 $\mu V$ RMS 0.16--3500 Hz),
5 kHz sampling rate (NeurOne, Mega Electronics Ltd, Kuopio, FI) (3.)
actively shielded EEG caps: 128 channels (WaveGuard Original, ANT,
Enschede, NL) and (4.) full optical decoupling: All devices are battery
powered and communicate via optic fibers.

Subjects sat in a comfortable armchair in the dimly lit Faraday cabin.
The contact impedance from electrodes to skin was typically reduced
below 5 kOhm using electrolyte gel (SUPER-VISC, EASYCAP GmbH,
Herrsching, GER) and blunt cannulas. Visual cues were presented using a
monitor outside the cabin, visible through the shielded window. The
distance between the display and the subjects' eyes was approx. 1 m. A
fixation point was attached at the center of the screen. The subjects
were instructed to relax, fixate the fixation mark and to keep as still
as possible during the motor execution task. Blinking and swallowing was
restricted to the inter-trial intervals.

The tasks were as following. Depending on the direction of a gray arrow
that was shown on black background, the subjects had to repetitively
clench their toes (downward arrow), perform sequential finger-tapping of
their left (leftward arrow) or right (rightward arrow) hand, or relax
(upward arrow). The movements were selected to require little
proximal muscular activity while still being complex enough to
keep subjects involved. Within the 4-s trials, the subjects performed
the repetitive movements at their own pace, which had to be maintained
as long as the arrow was showing. Per run, 80 arrows were displayed for
4 s each, with 3 to 4 s of continuous random inter-trial interval. The
order of presentation was pseudo-randomized, with all four arrows being
shown every four trials. Ideally 13 runs were performed to collect 260
trials of each movement and rest. The stimuli were presented and the
data recorded with BCI2000 \citep{schalk_bci2000:_2004}.
The experiment was approved by the ethical committee of
the University of Freiburg.

\subsection{EEG preprocessing}\label{subsec:supp-preprocessing}
We resampled the High-Gamma Dataset to 250 Hz, i.e., the same as the BCI Competition Dataset,
to be able to use the same ConvNet hyperparameter settings for both datasets.
To ensure that the ConvNets only have access to the same frequency range as the CSPs, we low-pass filtered the BCI 
Competition Dataset to below 38 Hz.
In case of the 4--$f_{end}$-Hz dataset, we highpass-filtered the signal as described in \ref{subsec:preprocessing} (
for the BCI Competition Dataset, we \emph{bandpass-filtered} to 4-38 Hz, so the previous lowpass-filter step was merged 
with the highpass-filter step).
Afterwards, for both sets, for the ConvNets, we performed electrode-wise exponential moving standardization with a decay
factor of 0.999 to compute  exponential moving means and variances for each channel and used these to standardize the 
continuous data.
Formally,
\begin{align}
x't& = (x_t - \mu_t) / \sqrt{\sigma_t^2} \\
\mu_t& = 0.001 x_t + 0.999\mu_{t-1} \\
\sigma_t^2& = 0.001(x_t - \mu_t)^2 + 0.999 \sigma_{t-1}^2
\end{align}

where $x't$ and $x_t$ are the standardized and the original signal for one electrode at time $t$, respectively.
As starting values for these recursive formulas we set the first 1000
mean values $\mu_t$ and first 1000 variance values $\sigma_t^2$ to the mean and the variance of the first 1000 samples,
which were always completely inside the training set (so we never used future test data in our preprocessing).
Some form of standardization is a commonly used procedure for ConvNets; exponentially moving standardization has
the advantage that it is also applicable for an online BCI.
For FBCSP, this standardization always worsened accuracies in preliminary experiments, so we did not use it.
We also did not use the standardization for our visualizations to ensure that the standardization does not make our
visualizations harder to interpret.
Overall, the minimal preprocessing without any manual feature extraction ensured our end-to-end pipeline could in 
principle 
be applied to a large number of brain-signal decoding tasks.

We also only minimally cleaned the data sets to remove extreme high-amplitude recording artifacts.
Our cleaning method thus only removed trials where at least one channel had a value outside
$\pm$800 $\mu V$.
We kept trials with lower-amplitude artifacts as we assumed these trials might still contain useful brain-signal 
information.
As described below, we used visualization of the features learned by the ConvNets to verify that they learned to 
classify brain signals and not artifacts.
Furthermore, for the High-Gamma Dataset, we used only those sensors covering the motor cortex: all central electrodes 
(45),
except the Cz electrode which served as the recording reference electrode.
Interestingly, using all electrodes led to worse accuracies for both the ConvNets and FBCSP, which may be a useful 
insight
for the design of future movement-related decoding/BCI studies.
Any further data restriction (trial-or channel-based cleaning) never led to accuracy increases in either of the two 
methods
when averaged across all subjects.
For the visualizations, we used all electrodes and common average re-referencing to investigate spatial distributions
for the entire scalp.

\subsection{Statistics}\label{subsec:statistics}

We used Wilcoxon signed-rank tests to check for statistical significance
of the mean difference of accuracies between decoding methods \citep{wilcoxon_individual_1945}.
We handled ties by  using the average rank of all tied data points and zero differences by assigning half of the 
rank-sum of these zero differences to the positive rank-sums and the other half to the negative rank-sums.
In case of a non-integer test-statistic value caused by ties or zeros, we rounded our test-statistic to the next 
larger integer, resulting in a more conservative estimate.

\subsection{Software implementation and hardware}\label{subsec:software-hardware}

We performed the ConvNet experiments on Geforce GTX Titan Black GPUs
with 6 GB memory. The machines had Intel(R) Xeon(R) E5-2650 v2 CPUs @
2.60 GHz with 32 cores (which were never fully used as most computations
were performed on the GPU) and 128 GB RAM. FBCSP was computed on an
Intel(R) Xeon(R) CPU E5-2650 v2 @ 2.60 GHz with 16 cores and 64 GB RAM.
We implemented our ConvNets using the Lasagne
framework \citep{dieleman_lasagne:_2015}, preprocessing of the data and FBCSP were implemented with
the Wyrm library \citep{venthur_wyrm:_2015}.
We will make the code to reproduce these results publicly
available.

\end{document}